**TITLE**: A Multi-Center Study on the Adaptability of a Shared Foundation Model for Electronic Health Records

**AUTHORS**: *[1]Lin Lawrence Guo,*[2]Jason Fries, [2]Ethan Steinberg, [2]Scott Lanyon Fleming, [3]Keith Morse, [4]Catherine Aftandilian, [5]Jose Posada, **[2]Nigam Shah, **[1,6]Lillian Sung

* Co-first authors; ** Co-senior authors

**AFFILIATIONS**:

[1]Program in Child Health Evaluative Sciences, The Hospital for Sick Children, Toronto, ON, Canada

[2]Stanford Center for Biomedical Informatics Research, Stanford University, Palo Alto, CA, USA

[3]Division of Pediatric Hospital Medicine, Department of Pediatrics, Stanford University, Palo Alto, CA, USA

[4]Division of Hematology/Oncology, Department of Pediatrics, Stanford University, Palo Alto, CA, USA

[5]Universidad del Norte, Barranquilla, Columbia

[6]Division of Haematology/Oncology, The Hospital for Sick Children, Toronto, ON, Canada

**ADDRESS FOR CORRESPONDANCE**:

Lillian Sung MD, PhD

Division of Haematology/Oncology

The Hospital for Sick Children,

555 University Avenue,

Toronto, Ontario, M5G1X8, Canada

Telephone: 416-813-5287

Fax: 416-813-5979

Email: Lillian.sung@sickkids.ca

WORD COUNT: Abstract 315; Text 3868; Tables: 3; Figures: 5; Appendices: 14

**Abstract**

**Background**: Foundation models hold promise for transforming artificial intelligence (AI) in healthcare by providing modular components that are easily adaptable to downstream healthcare tasks, making AI development more scalable and cost-effective. Foundation models for structured electronic health records (EHR), trained on coded medical records from millions of patients, demonstrated benefits including increased performance with fewer training labels, and improved robustness to distribution shifts. However, questions remain on the feasibility of sharing these models across different hospitals and their performance for local task adaptation.

**Objective**: This multi-center study examined the adaptability of a recently released structured EHR foundation model ($FM_{SM}$), trained on longitudinal medical record data from 2.57M Stanford Medicine patients.

**Methods**: Experiments were conducted using EHR data at The Hospital for Sick Children (SickKids) and Medical Information Mart for Intensive Care (MIMIC-IV). We assessed both adaptability via continued pretraining on local data, and task adaptability compared to baselines of training models from scratch at each site, including a local foundation model. We evaluated the discrimination performance and calibration of these models on 8 clinical prediction tasks.

**Results**: In both datasets, adapting the off-the-shelf $FM_{SM}$ matched the performance of gradient boosting machine (GBM) models locally trained on all available data while providing a 13% improvement in settings with few task-specific training labels. With continued pretraining on local data, label efficiency substantially improved, such that $FM_{SM}$ required fewer than 1% of training examples to match the fully trained GBM's performance. Continued pretraining was also 60 to 90% more sample-efficient than training local foundation models from scratch. Furthermore, our findings provided insights into when it is beneficial to adapt an existing EHR foundation model vs. pretraining from scratch, depending on data availability.

**Conclusion**: Our findings show that adapting shared EHR foundation models across hospitals provides improved prediction performance at less cost, underscoring the utility of base foundation models as modular components to streamline the development of healthcare AI.

## Introduction

*Foundation models*[1], large-scale artificial intelligence (AI) models trained on massive amounts of unlabeled data using self-supervised learning, mark a paradigm shift for healthcare AI by moving away from bespoke, single-purpose models to generalist and more easily adaptable medical AI[2]. Foundation models open new opportunities to improve diagnostic and predictive capabilities, enable proactive interventions and improve patient care using a range of modalities including natural language[3,4], imaging[5], genomics[6,7], and structured data from electronic health records (EHRs)[8-11]. Structured EHR foundation models, trained on tabular, timestamped event data for procedures, diagnoses, medications, and lab values as examples, offer distinct representational abilities over other modalities by focusing on encoding patients' longitudinal medical history. This enables generating feature representations that summarize a patient's entire medical history up to a specific time point, facilitating downstream tasks such as risk stratification and time-to-event modeling.

Recent EHR foundation models report state-of-the-art accuracy, require fewer labeled examples for task adaptation, and have demonstrated improved robustness to distribution shifts across time and patient subpopulations[12,13]. With model hubs (centralized repositories for pretrained model weights) playing a key role in modern AI development, sharing EHR foundation models across sites offers many practical advantages by providing a less expensive route for local hospitals to adapt a foundation model for their specific needs. More importantly, key properties of foundation models, such as their skills, domain knowledge, and biases, are highly dependent on the specific data used for pretraining[14,15]. Since large-scale EHR datasets (>1 million patients) are challenging to obtain for most researchers, sharing EHR foundation model weights becomes critical to advancing research into mitigating biases, improving robustness, and other properties intrinsic to a specific set of pretrained model weights. Finally, given recent arguments for regulatory oversight and quality assurance of healthcare AI models by public-private entities[16], access to foundation model weights that have undergone some certification process may become a prerequisite for model deployment.

Adapting and improving existing foundation models (rather than pretraining from scratch) is the predominant workflow in domains such as NLP and computer vision. However, the absence of public structured EHR foundation models has hampered similar progress in EHR settings[17]. This creates challenges in advancing label/sample efficiency, few-shot learning, and general methods to improving EHR foundation models without access to the original pretraining data[18]. For example, work in other modalities has found that pre-training on large-scale, heterogeneous data generally improves robustness[19] and that *continued pretraining* of existing models using in-domain data further improves performance in a target domain[20]. This offers a promising route to improving existing EHR foundation models at local hospitals but introduces potential challenges around catastrophic forgetting and other issues that have been underexplored due to the lack of large-scale, shared EHR models.

Although there is a growing body of work evaluating pretrained models across different hospital systems (GenHPF[21], TransformEHR[22]) and transfer from EHR data to insurance claims (Med-BERT[9]), prior studies have focused on private foundation models, pretrained from scratch, and the role architectural choices play in transfer learning performance in downstream task adaptation. There has been limited exploration of label efficiency in EHR settings, where encoder-only/BERT-style models perform poorly on few-shot tasks. For example, Med-BERT required an average of 200 to 1000 training examples per adapted task to outperform their reported logistic regression baselines. To our knowledge, no prior work has investigated strategies for improving existing EHR foundation models via continued pretraining or evaluations on how such training impacts label efficiency in downstream, adapted task models.

To address these challenges, we present a multi-center study focused on the adaptability of CLMBR-T-base[23], a recently released 141M parameter, decoder-only Transformer model for structured EHR data. This publicly accessible foundation model was pretrained from scratch on longitudinal, structured medical records from 2.57M patients from Stanford Medicine (FM$_{SM}$) and is compatible with the widely adopted Observational Medical Outcomes Partnership Common Data Model (OMOP CDM). The model's architecture has undergone extensive evaluation at Stanford Medicine across various settings[8,12,13,23].

**Our main contributions are summarized as follows:**

- The first multi-site evaluation of continued pretraining using a structured EHR foundation model. We evaluate cross-site, continued pretraining and adaptation for three foundation models trained from scratch and reflecting three different health systems: Stanford Medicine, SickKids (Toronto), and MIMIC-IV. We find continued pretraining improved performance in all models by an average of 3%.

- We evaluate the impact of continued pretraining on few-shot performance using 8 clinician-curated evaluation tasks and training sizes ranging from 2 to 1024 examples. We find substantial improvements to label efficiency, where on average 128 training examples can match baseline performance of a GBM trained on all available task data. This significantly outperforms prior label efficiency experiments that used encoder-only architectures.

- Continued pretraining results in foundation models that are similar in performance to those trained from scratch, but require 60 to 90% less patient data for pretraining.

## Methods

### Hospital Datasets

This retrospective multi-center study utilized EHR data collected from two institutions, The Hospital for Sick Children (SickKids) in Toronto, Canada, and Beth Israel Deaconess Medical Center (BIDMC) in Boston, United States.

The SickKids (SK) dataset was sourced from the SickKids Enterprise-wide Data in Azure Repository (SEDAR)[24] and contains EHR data for 1.8M patients collected from June 2018 to March 2023. The MIMIC dataset (MIMIC-IV version 1.0)[25,26] encompasses data from 340K patients admitted to an intensive care unit (ICU) or the emergency department at BIDMC between 2008 and 2019. To meet the schema requirements of the external foundation model, both datasets are mapped to the widely adopted Observational Medical Outcomes Partnership Common Data Model (OMOP-CDM). Mapping for MIMIC was done using code from the MIMIC project as part of Observational Health Data Sciences and Informatics[27].

The SickKids Research Ethics Board (REB) approved the use of SEDAR for this research (REB number: 1000074527). Data from MIMIC-IV was approved under the oversight of the Institutional Review Boards (IRB) of BIDMC and Massachusetts Institute of Technology (MIT). Data access was credentialed under the oversight of the data use agreement through PhysioNet[26].

**Models**

We evaluated the out-of-the-box performance of an external foundation model and its performance with continued pretraining. These results were compared against models that were locally trained from scratch, including baseline gradient boosted machines (GBMs) and a foundation model, as illustrated in Figure 1.

***Baseline GBMs – $GBM_{SK}$ and $GBM_{MIMIC}$.*** Our baseline approach employed a conventional count-based featurization approach[28,29] to create representations for each patient, task, and cohort, detailed further in Supplementary Table 1. Using these representations, we trained GBM models using LightGBM[30] on the task-specific training sets, resulting in $GBM_{SK}$ for SK and $GBM_{MIMIC}$ for MIMIC. We considered the choice of optimization algorithm as a hyperparameter.

Detailed hyperparameter settings are available in Supplementary Table 2, with tuning performed on the task-specific validation sets to optimize for log loss.

***External Foundation Model Stanford Medicine – FM<sub>SM</sub>.*** We selected CLMBR-T-base[23] as the external foundation model (FM<sub>SM</sub>), given it is the only publicly available EHR foundation model that has also undergone extensive evaluation[8,12,13,23]. FM<sub>SM</sub> processes sequences of clinical events as inputs, where each sequence $X = (x_1, x_2, \ldots, x_n)$ represents the medical timeline of an individual patient, with each $x_i$ denoting the *i*-th code, encompassing any form of structured data obtained from the patient's EHR, such as a diagnosis, procedure, medication, or lab test as examples. The architecture of the model comprises 12 stacked transformer layers with a local attention mechanism, totaling 141 million parameters. It was pretrained on structured EHRs from 2.57 million patients receiving care at Stanford Health Care and Lucile Packard Children's Hospital between 2008 and 2022. CLMBR-T-base is a decoder-only model, autoregressively pretrained to predict the next clinical event $x_{t+1}$ based on the preceding sequence, similar to GPT pretraining. The vocabulary is defined as the top 65,536 codes from the union of all codes in 21 source ontology mappings (e.g. LOINC, SNOMED, RxNorm) provided by Athena's OMOP vocabulary list. The top codes were selected based on global code frequency in the original pretraining dataset. Codes not in the vocabulary were dropped.

In our experiments, we used FM<sub>SM</sub> out-of-the-box as a frozen feature encoder, transforming the sequence of clinical events *X* for each patient into a dense vector representation $R = f_\theta(X)$, with the model's parameters $\theta$ fixed during this encoding process. The obtained representations were then used to train linear task heads (also known as linear probes[31]) on the task-specific training sets. Specifically, we utilized L2-regularized logistic regression models from Sci-kit Learn[32] with the LBFGS solver as our linear task heads. We conducted hyperparameter tuning on the task-specific validation sets (see Supplementary Table 2 for hyperparameter settings) to optimize for log loss.

***External Foundation Model Stanford Medicine with Continued Pretraining – $FM_{SM}^{+SK}$ and $FM_{SM}^{+MIMIC}$***. We tailored $FM_{SM}$ to each dataset separately by performing continued pretraining, using patient timelines from each dataset, resulting in $FM_{SM}^{+SK}$ and $FM_{SM}^{+MIMIC}$. Continued pretraining uses the same next-code, autoregressive prediction task used during the original pretraining run. Pretraining resumed on the global training set, tuning the learning rate using global validation set performance. Pretraining continued for up to 10 million steps, with early stopping implemented if the validation loss did not improve for 15,000 steps. Once continued pretraining was complete, the training of linear task heads followed the same procedure as $FM_{SM}$.

***Local Foundation Models – $FM_{SK}$ and $FM_{MIMIC}$.*** The local foundation models were pretrained from scratch on each dataset using up to 1.6M patients (100M coded events) for SK and 290K (180M coded events) for MIMIC, resulting in $FM_{SK}$ and $FM_{MIMIC}$. $FM_{SK}$ and $FM_{MIMIC}$ shared the same architecture as $FM_{SM}$. Hyperparameter tuning followed the same procedure as continued pretraining.

**Clinical Prediction Tasks, Prediction Times, and Observation Window**

We defined 8 binary clinical prediction tasks. The binary classification setting is standard and particularly well-suited for the class of models being evaluated, and the tasks were selected based on clinical relevance, alignment with previous benchmarks[13,23], and previous validation work[33]. The tasks were categorized into two groups: operational tasks and predicting clinically relevant abnormal lab results. Operational tasks encompassed in-hospital mortality, long length of stay of at least 7 days (long LOS), readmission within 30 days of discharge (30-day readmission). Tasks related to predicting abnormal lab-based results included hypoglycemia

(serum glucose <3 mmol/L) [34], hyponatremia (serum sodium <125 mmol/L)[35], hyperkalemia (serum potassium >7 mmol/L)[36], anemia (hemoglobin <70 g/L)[37], and thrombocytopenia (platelet count < 50 x $10^9$/L).

The prediction time for the 30-day readmission task was set at midnight on the day of discharge. For all other tasks, the prediction time was set at midnight on the day of admission. The prediction window extended until discharge for all tasks, with the exception of the readmission task, which utilized a 30-day window post-discharge. The observation window for model input encompassed all available patient data up to the prediction time.

**Inpatient Cohort and Data Splitting Procedure**

Supplementary Figure 1 illustrates the assignment of patients into an inpatient cohort. SickKids inpatients were those 28 days of age or older at admission. MIMIC inpatients were those 18 years of age or older at admission. In cases where patients had multiple admissions, we randomly selected one admission for inclusion. We excluded admissions in which patient death or discharge occurred between admission and prediction time. Additionally, for each clinical prediction task, we excluded admissions in which the outcome occurred between admission (or discharge) and prediction time.

We defined global training, validation, and test sets for each dataset in a 70/15/15 ratio across all patients used for continued pretraining and local foundation models. Subsequently, for each inpatient cohort, we derived task-specific training, validation, and test sets. The task-specific sets were subsets of the global sets, thereby preserving the 70/15/15 distribution.

**Computation Resources**

Pretraining was executed on an on-premises cluster of up to 4 Nvidia V100 GPUs.

**Experiments**

***Overall Performance***. We compared discrimination (AUROC) and calibration (expected calibration error, ECE) of $FM_{SM}$ and $FM_{SM}$ with continued pretraining ($FM_{SM}^{+}$) vs. the baseline GBM across all clinical prediction tasks using the task-specific test set of each dataset. We also compared discrimination and calibration of $FM_{SM}$ and $FM_{SM}^{+}$ with local foundation models ($FM_{SK}$ and $FM_{MIMIC}$). In an ablation experiment, we evaluated the adaptability of $FM_{SM}$ against $FM_{SK}$ and $FM_{MIMIC}$, hypothesizing that $FM_{SM}$'s training on a larger and more heterogenous patient population makes it more adaptable across healthcare sites. To this end, we examined the performance of $FM_{SK}$ on MIMIC and $FM_{MIMIC}$ on SK as external foundation models, including their performance with continued pretraining ($FM_{SK}^{+MIMIC}$ and $FM_{MIMIC}^{+SK}$), compared to $FM_{SM}$ and $FM_{SM}^{+}$.

***Few-Shot Performance.*** We compared $FM_{SM}$ and continued pretraining of $FM_{SM}$ ($FM_{SM}^{+}$) against the baseline GBM in the setting of decreasing training samples to evaluate label efficiency. For each task, we varied the number of training examples (*k*) within (2, 4, 8, 16, 32, 64, 128, 256, 512, 1024). These examples consisted of an equal number of positive and negative instances, except for tasks with $k_p$ positive instances, where $k_p$ < *k/2*. For such tasks, we included all $k_p$ positive instances in our training data. The training examples were drawn from a subset of the task-specific training set, while validation and test sets remained the same. We do not k-sample the validation set under the assumption that hospitals, in practice, will have access to larger task sets for evaluating model performance. In order to ensure an unbiased performance estimate, we removed all task-specific training sets from pretraining for these experiments to ensure that pretraining is not unfairly advantaged by being able to see more examples of a particular task. The procedures for training baseline GBM and linear task heads for foundation models were kept consistent for the few-shot experiments. We performed 20 iterations of sampling, task-specific training, and evaluation for each value of *k*. Notably, our method of training linear task heads on $FM_{SM}$ and $FM_{SM}^{+}$ across all experiments does not involve fine-tuning the foundation model parameters, representing a conservative approach. In contrast, the baseline GBMs underwent extensive hyperparameter tuning, including adjustments to the optimization algorithm, to optimize performance.

***Performance with Varying Pretraining Sample Size.*** We evaluated the effect of decreasing pretraining sample size on the performance of continued pretraining ($FM_{SM}^{+SK}$ and $FM_{SM}^{+MIMIC}$) and pretraining local foundation models from scratch ($FM_{SK}$ and $FM_{MIMIC}$). Our aims were twofold: 1) to provide insight into the decision of whether to employ or tailor (via continued pretraining) $FM_{SM}$ versus pretraining from scratch, depending on sample size availability; and 2) to determine if continued pretraining is more sample efficient than pretraining from scratch. For the first aim, we compared the performance of $FM_{SM}$ with and without continued pretraining against local FMs at each sample size. For the second aim, we assessed how continued pretraining at various sample sizes compared to local FMs trained on the entire dataset. The subsamples used varied from 0.1% to 80% of the global training set, for both continued pretraining and pretraining from scratch.

**Model Evaluation and Statistical Analysis**

We evaluated each model's discrimination performance in the task-specific test sets using the AUROC. Calibration was measured in terms of the ECE using 10 quantile bins for the predicted risks. To calculate ECE in the few-shots experiment, we determined the corrected predicted risk *p'* based on the predicted risk *p* and the task-specific outcome rate *b'* using Formula 1. The correction accounts for biases in the model's predictions as a result of balanced sampling used during few-shot training[38].

Formula 1. $$p' = \frac{p}{p + (1-p)(1-b')/b'}$$

Differences between model performances were statistically evaluated using a hierarchical bootstrapping method, which samples both patients and outcomes with replacement[39]. In the few-shots experiments, we averaged the performance across all sampling iterations. For each statistical evaluation, we calculated a two-tailed *P*-value by multiplying the one-sided *P*-value by 2, where the one-sided *P*-value represents the smaller proportion of bootstrap differences either to the left of 0 (the null value) or to the right of 0[40]. For all tests, a *P*-value of <0.05 was considered statistically significant.

## Results

Table 1 presents patient demographic characteristics and outcome prevalence within the SickKids and MIMIC inpatient cohorts. Outcome prevalence across the two cohorts was highly skewed, with MIMIC having higher rates in 6/8 tasks (1.3 to 6 times larger than SickKids).

Table 2 shows comparisons of mean AUROC for $FM_{SM}$ and $FM_{SM}$ with continued pretraining ($FM_{SM}^+$) vs. the baseline GBMs. Supplementary Table 3 shows comparisons against local FMs. $FM_{SM}^+$ had significantly better AUROC compared to GBM in both SickKids and MIMIC cohorts. ECE was also significantly better for $FM_{SM}$ and $FM_{SM}^+$ vs. GBMs in both cohorts. In the ablation experiment focused on the adaptability of external foundation models, Figure 2 demonstrates that $FM_{SM}$ and $FM_{SM}^+$ achieved significantly higher AUROC compared to $FM_{SK}$ and $FM_{SK}^+$ on the MIMIC dataset, and to $FM_{MIMIC}$ and $FM_{MIMIC}^+$ on the SK dataset. See Table 3 for per-task AUROC and Supplementary Table 4 for per-task ECE.

In the few-shot settings (Figure 3 and Supplementary Table 5), $FM_{SM}$ and $FM_{SM}^+$ consistently demonstrated significantly improved AUROC against GBM across almost all few-shot settings in both datasets (13% improvement on average for $FM_{SM}$ and 19% for $FM_{SM}^+$).

Furthermore, $FM_{SM}^{+}$ matched mean AUROC of GBM using as little as 128 samples (64 samples per class; fewer than 1% of all training examples). Supplementary Figure 2 and Supplementary Table 6 show that $FM_{SM}$ and $FM_{SM}^{+}$ demonstrated lower ECE against GBM although the difference was not statistically significant in 512- and 1024-shot settings, and in the 32-shot setting for SK.

In the experiments with varying pretraining sample size (Figure 4 and Supplementary Table 7), AUROCs for external foundation models ($FM_{SM}$ and $FM_{SM}^{+}$) were significantly better than local foundation models with very small subsamples (up to 1% for SickKids and 0.1% for MIMIC). Without continued pretraining, AUROC for $FM_{SK}$ was significantly better than $FM_{SM}$ when subsamples reached 80% for SickKids. In contrast, AUROC for local FMs and $FM_{SM}^{+}$ did not differ significantly across larger subsamples. Calibration results, detailed in Supplementary Figure 3 and Supplementary Table 8, did not significantly vary between models.

Figure 5 and Supplementary Table 9 show that the performance of continued pretraining with subsamples of 10% or more for SK and 40% or more for MIMIC did not significantly differ from that of local foundation models ($FM_{SK}$ and $FM_{MIMIC}$) trained from scratch on all data.

In addition, continued pretraining was more efficient than pretraining from scratch (70.2% faster than $FM_{SK}$ and 58.4% faster than $FM_{MIMIC}$), as shown in Supplementary Figure 4. Furthermore, as demonstrated in Supplementary Figure 5, $FM_{SM}$ and $FM_{SM}^{+}$ generally processed fewer coded events (50.9% less in SK and 30.9% less in MIMIC) than local foundation models as a result of reduced code coverage.

## Discussion

Our multi-center study has demonstrated that adapting an off-the-shelf external foundation model ($FM_{SM}$) can yield comparable discrimination and better calibration compared to baseline GBM models locally trained on all available data at each site, while providing 13% discrimination improvement in settings with few task-specific training labels. With continued pretraining on local data, significantly better discrimination and calibration were observed compared to baseline GBM models. In addition, label efficiency substantially improved, such that using only 128 training examples (64 samples per class; less than 1% of the total data), $FM_{SM}^{+}$ matched the performance of the GBMs training using all available data. Furthermore, continued pretraining of $FM_{SM}$ requires 60 to 90% less patient data than local pretraining from scratch to achieve the same level of performance, demonstrating better sample efficiency. Our findings also provided insights into when it is beneficial to adapt an existing EHR foundation model vs. pretraining from scratch, depending on data availability.

The development of single-purpose models entails significant costs[41]. These costs escalate further for EHR foundation models, which require access to extensive patient records and substantial computing resources. Our findings highlight the potential for cost savings by sharing and building upon pretrained EHR foundation models across hospitals. Adapting these models to new tasks significantly reduces the amount of training labels needed, thereby lowering label acquisition costs and speeding up the deployment of new applications. Moreover, for institutions equipped to conduct pretraining, the continued pretraining of an existing foundation model is considerably more data- and compute-efficient compared to pretraining from scratch.

Our results contribute to recent work demonstrating the robustness of EHR foundation models to various distribution shifts[12,13]. Compared to traditional single-purpose models that often fall short in adaptability across sites[42,43], EHR foundation models demonstrate a notable capacity to encode patients' longitudinal medical history and local care nuances. Despite challenges posed by dataset shifts, notably in patient populations and coding variations as evidenced by the number of unprocessed codes by $FM_{SM}$ across all patient timelines at each site, the external foundation model displayed robust performance. Moreover, our results indicate that pretraining on a larger and more diverse patient population improves the adaptability of the foundation model across healthcare settings. It is noteworthy that a single external foundation model consistently achieved strong performance across both a Canadian pediatric cohort and an American adult ICU-based cohort. Our findings point towards a paradigm where, instead of training bespoke models for each healthcare site from scratch, the focus shifts to the development and sharing of larger, general-purpose base foundation models and recipes for site-specific refinement, such as continued pretraining.

To leverage the transfer learning capabilities of pretrained foundation models, it is important to adhere to the underlying data schema and vocabulary of tokens used by these models. For this reason, we mapped our datasets to the OMOP CDM. Progress is required to define minimal schema requirements for training foundation models to mitigate costs associated with mapping to a common data model. Alternatively, developing EHR foundation models that are robust to schema shifts[44], akin to multilingual language models, represents a valuable direction for future research.

Sharing foundation models across hospitals needs to address privacy and ethical use of the underlying data. For EHR foundation models, adopting measures such as training on data de-identified to HIPAA standards, acquiring patient consent, and restricting access to accredited individuals[11] are important first steps. Despite these efforts, issues like potential misuse (e.g., for surveillance) and the challenge of algorithmic biases remain open research questions[1,45], underscoring the ongoing challenges in maintaining privacy and mitigating risks in medical foundation models.

This study has several strengths and limitations. A primary strength of this study is that it represents one of the first external evaluation of a publicly available foundation model specifically for structured EHR data. Additionally, we have characterized the utility of the foundation model across diverse evaluation settings. On the limitation side, this study was conducted using a limited number of hospital datasets and tasks, which may not capture the full spectrum of EHR variability. We did not explore methods for harmonizing across schemas, which could impact the adaptability of the foundation model. We do not explore trade-offs of relative set sizes of training vs validation data when conducting few-shot learning, which may impact reported performance. We also did not explore questions on fairness and the propagation of biases potentially associated with sharing foundation models. Moreover, the study emphasized a specific foundation model, CLMBR-T-base, due to lack of publicly available structured EHR foundation models[17]. Lastly, while the benefits of continued pretraining are clear, they may not be accessible to institutions that lack resources to perform it, potentially limiting broader applicability.

## Conclusions

Our findings show that adapting a pretrained EHR foundation model for downstream tasks across hospitals can improve prediction performance at less cost, underscoring the effectiveness of sharing base foundation models as modular machine learning components to streamline the development of healthcare AI.

## Data Availability

The SickKids dataset cannot be made publicly available because of the potential risk to patient privacy. The MIMIC-IV dataset is available from https://mimic.mit.edu/docs/iv/. Details including how to access the external foundation model used in this study are available from https://ehrshot.stanford.edu/.

## Code Availability

The code for the analyses conducted in this study is open-source and available at https://github.com/sungresearch/femr-on-sk and https://github.com/sungresearch/femr-on-mimic.

## Acknowledgments

We would like to acknowledge Joshua Lemmon and Jiro Inoue for their contributions to the curation of the datasets used in this study. We would like to also thank Stephen R. Pfohl for his contributions to model evaluation. LS is supported by the Canada Research Chair in Pediatric Oncology Supportive Care.

## Competing interests

The authors declare that they have no competing interests.

## Funding

No external funding was received for the study.

## Authors' contributions

LLG, JF, and LS conceptualized and designed the study with input from all authors. LLG performed all experiments. ES contributed to the codebase. LLG, JF, and LS analyzed and interpreted results with input from all authors. LLG and JF wrote the manuscript. All authors revised and commented on the manuscript. All authors read and approved the final manuscript.

Table 1. Cohort characteristics and outcome prevalence

|  | SickKids (n = 37,960) | MIMIC (n = 44,055) |
|---|---|---|
| Median age [IQR] | 7 [2, 13] | 56 [35, 71] |
| Sex, n (%) | | |
|   Male | 20,507 (54.0%) | 17,329 (39.3%) |
| Race, n (%) | | |
|   White | ** | 27,402 (62.2%) |
|   Black or African American | ** | 5,338 (12.1%) |
|   Asian | ** | 1,799 (4.1%) |
|   Other | ** | 5,066 (11.5%) |
|   Unknown or Unable to Obtain | ** | 4,450 (10.1%) |
| *Outcomes, n (%) | | |
|   In-hospital Mortality | 216 (0.6%) | 1,599 (3.6%) |
|   Long LOS | 6,115 (16.1%) | 12,215 (27.7%) |
|   30-day Readmission | 2,275 (6.0%) | 259 (0.6%) |
|   Hypoglycemia | 459 (1.2%) | 719 (1.6%) |
|   Hyponatremia | 92 (0.2%) | 385 (0.9%) |
|   Hyperkalemia | 352 (0.9%) | 375 (0.9%) |
|   Thrombocytopenia | 726 (1.9%) | 1,342 (3.1%) |
|   Anemia | 1,073 (2.9%) | 2,954 (6.8%) |

* Refer to text for outcome definitions
** Race data is not routinely collected at SickKids
Abbreviations: IQR: interquartile range; LOS: length of stay; MIMIC: Medical Information Mart for Intensive Care

Table 2. Comparing discrimination and calibration of foundation models vs. baseline GBM approaches at each site*

|  | Discrimination Evaluation | | | Calibration Evaluation | | |
|---|---|---|---|---|---|---|
| Model | Mean AUROC | Difference [Foundation Model – GBM] | P-value** | Mean ECE | Difference [Foundation Model - GBM] | P-value** |
| **SickKids** | | | | | | |
| $GBM_{SK}$ | 0.855 [0.798, 0.904] | | | 0.015 [0.013, 0.017] | | |
| $FM_{SM}$ | 0.880 [0.826, 0.928] | 0.024 [-0.009, 0.072] | 0.184 | 0.005 [0.003, 0.009] | -0.010 [-0.012, -0.006] | **< 0.001** |
| $FM_{SM}^{+SK}$ | 0.901 [0.851, 0.944] | 0.044 [0.013, 0.099] | **0.002** | 0.006 [0.003, 0.009] | -0.010 [-0.013, -0.006] | **< 0.001** |
| | | | | | | |
| **MIMIC** | | | | | | |
| $GBM_{MIMIC}$ | 0.807 [0.733, 0.868] | | | 0.016 [0.015, 0.018] | | |
| $FM_{SM}$ | 0.828 [0.775, 0.875] | 0.020 [-0.010, 0.062] | 0.230 | 0.007 [0.004, 0.012] | -0.010 [-0.013, -0.005] | **0.002** |
| $FM_{SM}^{+MIMIC}$ | 0.848 [0.783, 0.895] | 0.039 [0.009, 0.075] | **0.006** | 0.005 [0.003, 0.009] | -0.011 [-0.013, -0.008] | **< 0.001** |

* Table shows mean AUROC and ECE across tasks [95% hierarchical bootstrap CI]
** Bolded values indicate P < 0.05
Abbreviations: AUROC: area under the receiver operating characteristics curve; CI: confidence interval; $FM_{SM}$: external foundation model Stanford Medicine; $FM_{SM}^{+}$: external foundation model Stanford Medicine with continued pretraining - SK or MIMIC; SK: SickKids; MIMIC: Medical Information Mart for Intensive Care; GBM: gradient boosting machines.

Table 3. Discrimination for task-specific models at two sites*

| Model | In-hospital Mortality | Long LOS | 30-day Readmission | Hypoglycemia | Hyponatremia | Hyperkalemia | Thrombocytopenia | Anemia |
|---|---|---|---|---|---|---|---|---|
| Dataset: SickKids | | | | | | | | |
| $GBM_{SK}$ | 0.893 [0.815, 0.953] | **0.866 [0.853, 0.879]** | 0.783 [0.755, 0.809] | 0.880 [0.830, 0.924] | 0.783 [0.579, 0.957] | 0.749 [0.687, 0.808] | 0.953 [0.928, 0.975] | 0.919 [0.898, 0.938] |
| $FM_{MIMIC}$ | 0.885 [0.814, 0.943] | 0.797 [0.782, 0.813] | 0.752 [0.724, 0.779] | 0.899 [0.866, 0.927] | 0.936 [0.889, 0.972] | 0.716 [0.655, 0.773] | 0.930 [0.900, 0.956] | 0.895 [0.870, 0.918] |
| $FM_{MIMIC}^{+SK}$ | 0.946 [0.912, 0.973] | 0.830 [0.816, 0.844] | 0.798 [0.773, 0.821] | **0.927 [0.901, 0.950]** | 0.890 [0.789, 0.969] | 0.813 [0.767, 0.855] | 0.967 [0.953, 0.979] | 0.915 [0.892, 0.935] |
| $FM_{SM}$ | 0.941 [0.902, 0.971] | 0.815 [0.800, 0.830] | 0.774 [0.747, 0.799] | 0.915 [0.879, 0.945] | 0.925 [0.880, 0.963] | 0.793 [0.743, 0.838] | 0.962 [0.946, 0.975] | 0.918 [0.897, 0.937] |
| $FM_{SM}^{+SK}$ | 0.957 [0.922, 0.980] | 0.839 [0.825, 0.853] | **0.804 [0.780, 0.828]** | 0.918 [0.883, 0.948] | **0.957 [0.933, 0.978]** | 0.822 [0.781, 0.860] | **0.967 [0.950, 0.980]** | **0.946 [0.932, 0.959]** |
| $FM_{SK}$ | **0.968 [0.952, 0.981]** | 0.847 [0.834, 0.861] | 0.774 [0.746, 0.799] | 0.924 [0.891, 0.953] | 0.936 [0.882, 0.977] | **0.848 [0.812, 0.883]** | 0.966 [0.953, 0.978] | 0.932 [0.912, 0.950] |
| Dataset: MIMIC | | | | | | | | |
| $GBM_{MIMIC}$ | 0.905 [0.886, 0.922] | **0.831 [0.821, 0.841]** | 0.619 [0.514, 0.719] | 0.790 [0.746, 0.830] | 0.798 [0.723, 0.863] | 0.700 [0.619, 0.775] | 0.915 [0.884, 0.943] | 0.878 [0.862, 0.893] |
| $FM_{SK}$ | 0.874 [0.852, 0.894] | 0.753 [0.741, 0.764] | 0.596 [0.49, 0.693] | 0.768 [0.728, 0.805] | 0.811 [0.762, 0.857] | 0.728 [0.657, 0.795] | 0.918 [0.896, 0.939] | 0.805 [0.786, 0.824] |
| $FM_{SK}^{+MIMIC}$ | 0.920 [0.904, 0.934] | 0.792 [0.781, 0.802] | 0.614 [0.507, 0.716] | 0.808 [0.774, 0.840] | 0.846 [0.801, 0.888] | 0.782 [0.726, 0.836] | 0.946 [0.929, 0.962] | 0.854 [0.837, 0.869] |
| $FM_{SM}$ | 0.911 [0.894, 0.926] | 0.792 [0.781, 0.803] | **0.695 [0.601, 0.781]** | 0.812 [0.779, 0.844] | 0.817 [0.771, 0.859] | 0.800 [0.739, 0.856] | 0.928 [0.909, 0.946] | 0.854 [0.836, 0.871] |
| $FM_{SM}^{+MIMIC}$ | 0.927 [0.912, 0.941] | 0.823 [0.813, 0.833] | 0.673 [0.564, 0.771] | 0.837 [0.802, 0.869] | **0.845 [0.806, 0.882]** | **0.818 [0.762, 0.868]** | **0.952 [0.939, 0.964]** | 0.875 [0.859, 0.890] |
| $FM_{MIMIC}$ | **0.93 [0.916, 0.944]** | 0.826 [0.816, 0.836] | 0.690 [0.589, 0.784] | **0.838 [0.806, 0.869]** | 0.845 [0.799, 0.887] | 0.814 [0.763, 0.862] | 0.950 [0.934, 0.964] | **0.887 [0.872, 0.900]** |

*Table shows AUROC (95% bootstrap CI) for each task; bolded values indicate highest AUROC across models

Abbreviations: AUROC: area under the receiver operating characteristic curve; GBM: gradient boosting machines; $FM_{SM}$/$FM_{SK}$/$FM_{MIMIC}$: foundation model Stanford Medicine, SickKids, or MIMIC; $FM_{SM}^+$/$FM_{SK}^+$/$FM_{MIMIC}^+$: external foundation model with continued pretraining - SK or MIMIC; SK: SickKids; MIMIC: Medical Information Mart for Intensive Care; LOS: length of stay; CI: confidence interval

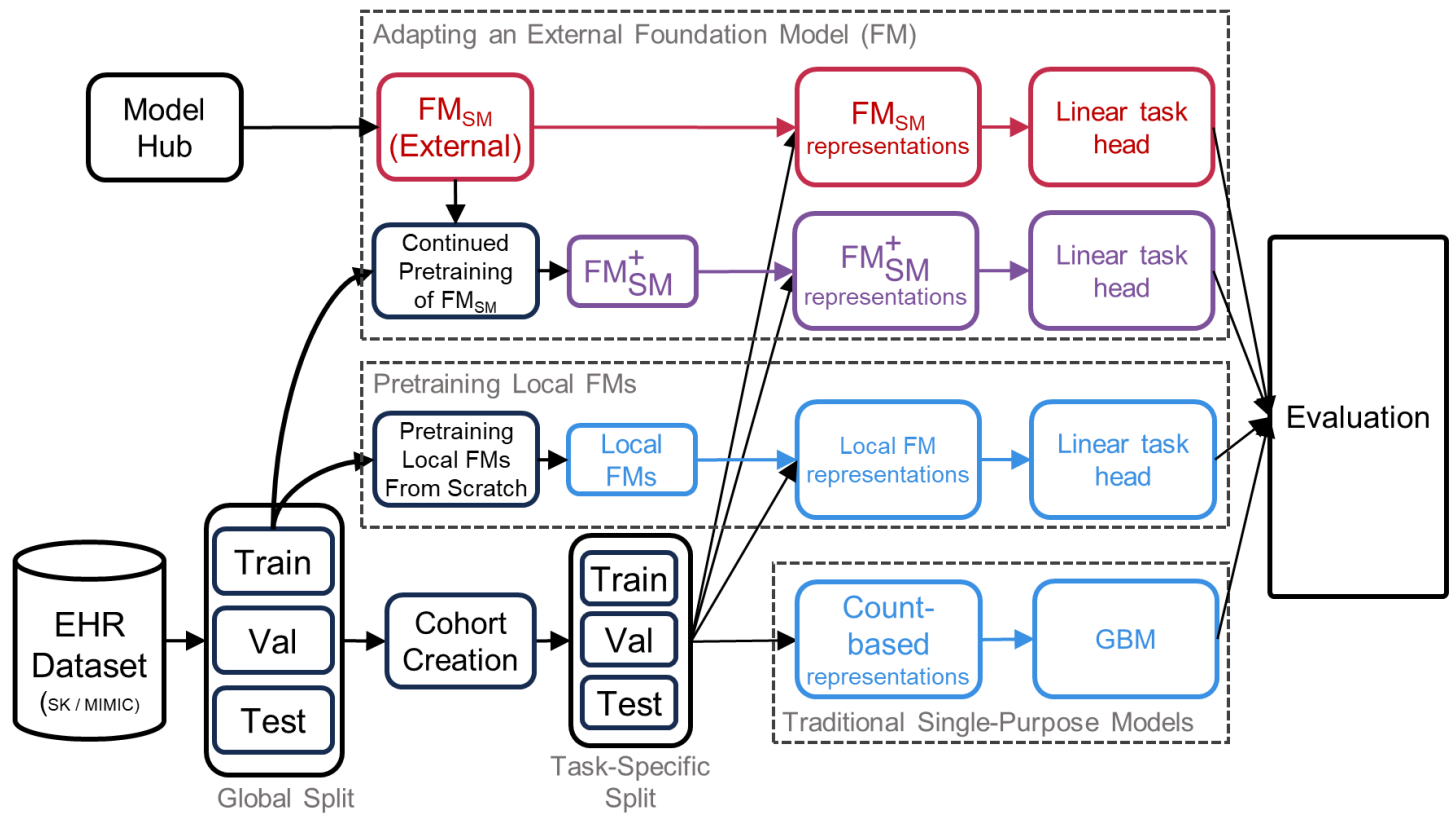

Figure 1. Overview of model training and evaluation. Patients in each dataset (SK and MIMIC) were globally split into training, validation, and test sets. An inpatient cohort was defined for patients in each dataset. We adopted an external FM ($FM_{SM}$), CLMBR-T-base, pretrained on structured EHRs of 2.57M patients from Stanford Medicine and used it to generate representations for each patient in the cohorts. We also conducted continued pretraining using $FM_{SM}$ on the global training set of each dataset and subsequently constructed patient representations using the resulting models ($FM_{SM}^{+SK}$ and $FM_{SM}^{+MIMIC}$). With these representations, we trained linear task heads (logistic regression) and compared them to locally trained models across the SK and MIMIC datasets and 8 evaluation tasks spanning operational outcomes and anticipated abnormal lab results.

Abbreviations: EHR: electronic health records; SK: SickKids; MIMIC: Medical Information Mart for Intensive Care; FM: foundation model; $FM_{SM}$: external foundation model Stanford Medicine; CLMBR-T: clinical language model based representation – transformer; GBM: gradient boosting machines; Val: validation.

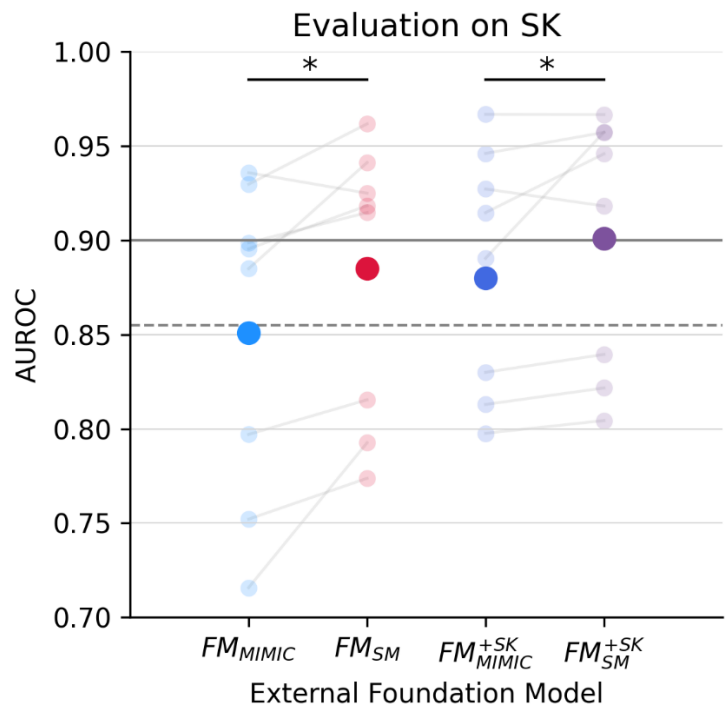 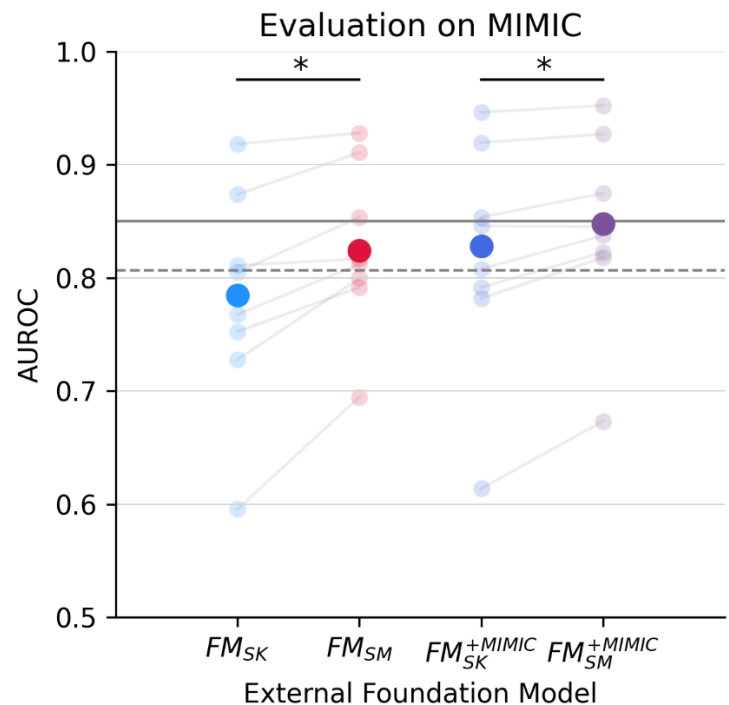

Figure 2. Comparing discrimination performance of external foundation models: $FM_{SM}$ vs. $FM_{MIMIC}$ on SK, and $FM_{SM}$ vs. $FM_{SK}$ on MIMIC. Grey lines indicate the performance of locally trained models, with the dashed lines indicating baseline GBMs and solid lines indicating the local foundation models. Bolded and faint dots indicate average and task-specific performance, respectively. Asterisks indicate significant differences at p<0.05 evaluated using hierarchical bootstrapping.

Abbreviations: AUROC: area under the receiver operating characteristic curve; $FM_{SM}$/$FM_{SK}$/$FM_{MIMIC}$: external foundation model Stanford Medicine, SK, or MIMIC; $FM^+_{SM}$/$FM^+_{SK}$/$FM^+_{MIMIC}$: external foundation model with continued pretraining – SK or MIMIC; SM: Stanford Medicine; SK: SickKids; MIMIC: Medical Information Mart for Intensive Care.

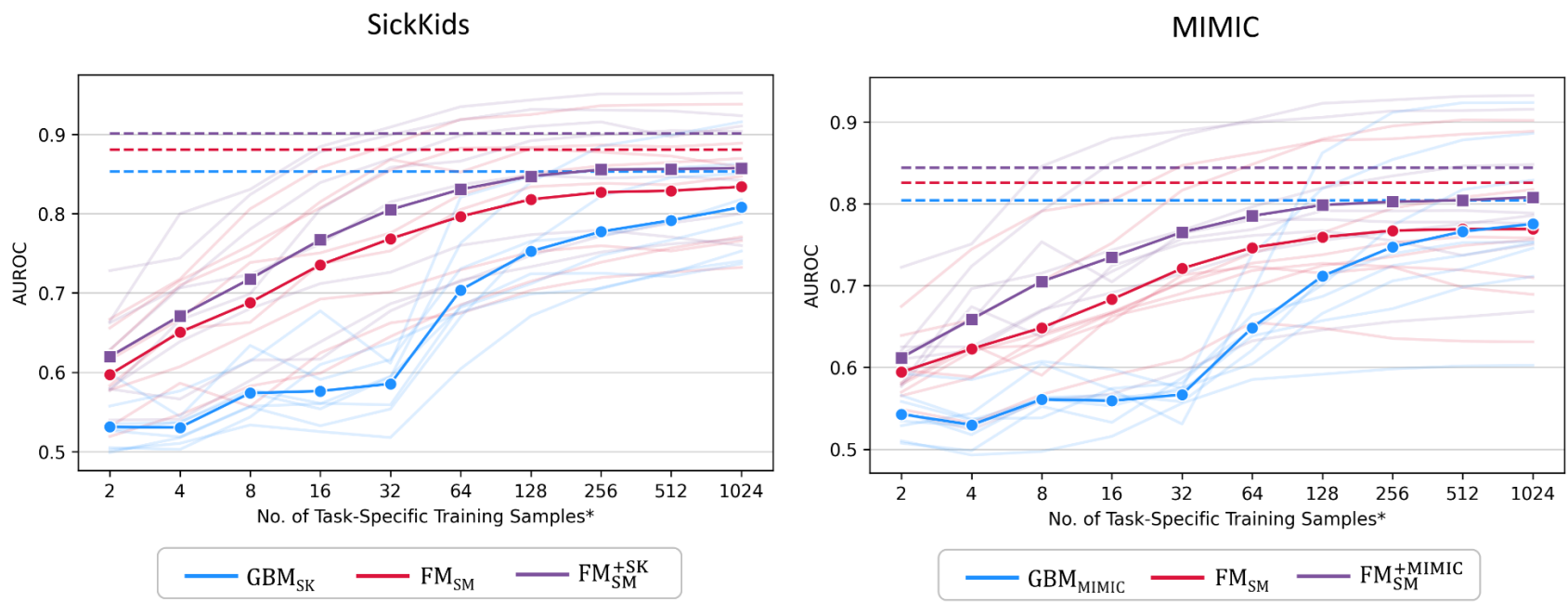

Figure 3. Discrimination of external foundation model ($FM_{SM}$), external foundation model with continued pretraining ($FM_{SM}^{+}$) and baseline GBM using decreasing training samples at SickKids and MIMIC. Bolded and faint lines indicate average and task-specific performance, respectively. Dashed lines indicate mean AUROC of models trained on all training samples.

* The number of training examples for each class is up to half of the number of task-specific training samples.

Abbreviations: AUROC: area under the receiver operating characteristic curve; $FM_{SM}$: external foundation model Stanford Medicine; $FM_{SM}^{+}$: external foundation model Stanford Medicine with continued pretraining – SK or MIMIC; SK: SickKids; MIMIC: Medical Information Mart for Intensive Care; GBM: gradient boosting machines.

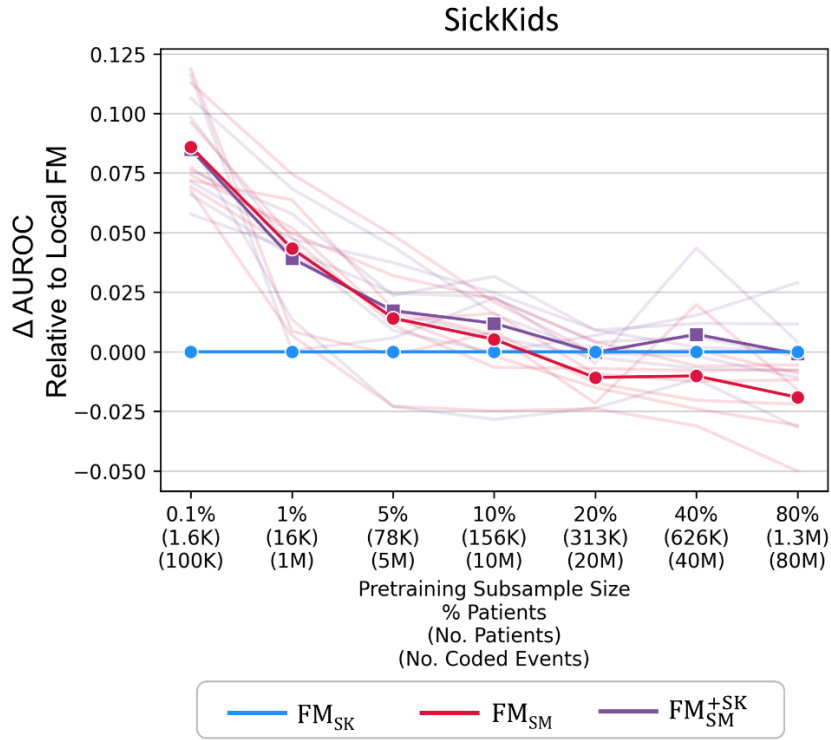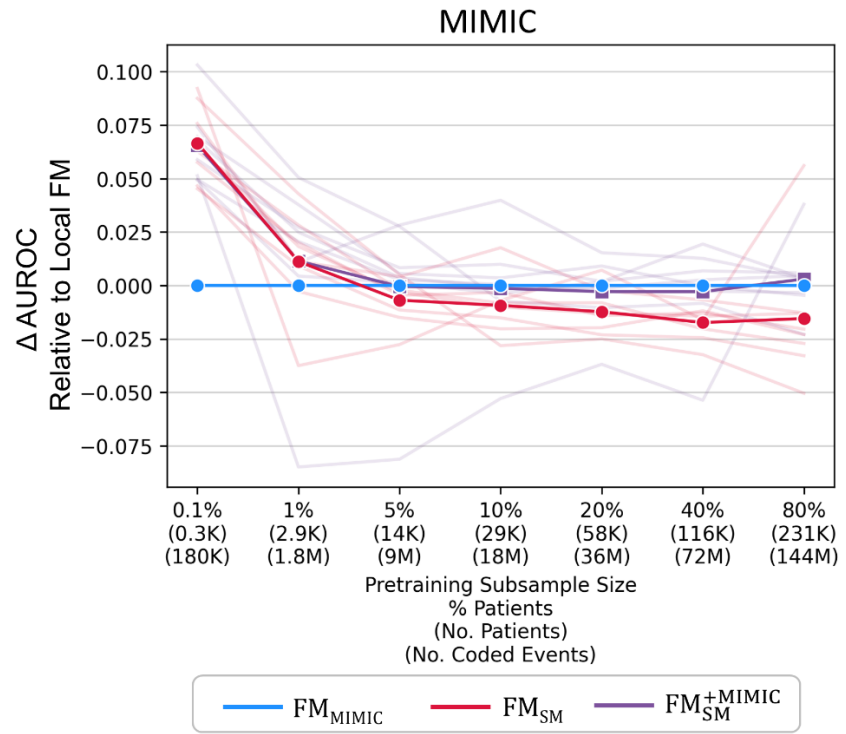

Figure 4. Discrimination of external foundation model ($FM_{SM}$) and external foundation model with continued pretraining ($FM_{SM}^{+}$) relative to local foundation model ($FM_{SK}$ and $FM_{MIMIC}$) using decreasing pretraining sample size. Bolded and faint lines indicate average and task-specific performance relative to local foundation models, respectively. The subsample size is not relevant for $FM_{SM}$, which did not undergo additional pretraining. Note, the model AUROC scores are relative to the baseline local models ($FM_{SK}$ and $FM_{MIMIC}$) and the absolute AUROC does change across sample sizes.

Abbreviations: AUROC: area under the receiver operating characteristic curve; $FM_{SM}$: external foundation model Stanford Medicine; $FM_{SM}^{+}$: external foundation model Stanford Medicine with continued pretraining – SK or MIMIC; $FM_{SK}/FM_{MIMIC}$: local foundation models – SK or MIMIC; SK: SickKids; MIMIC: Medical Information Mart for Intensive Care.

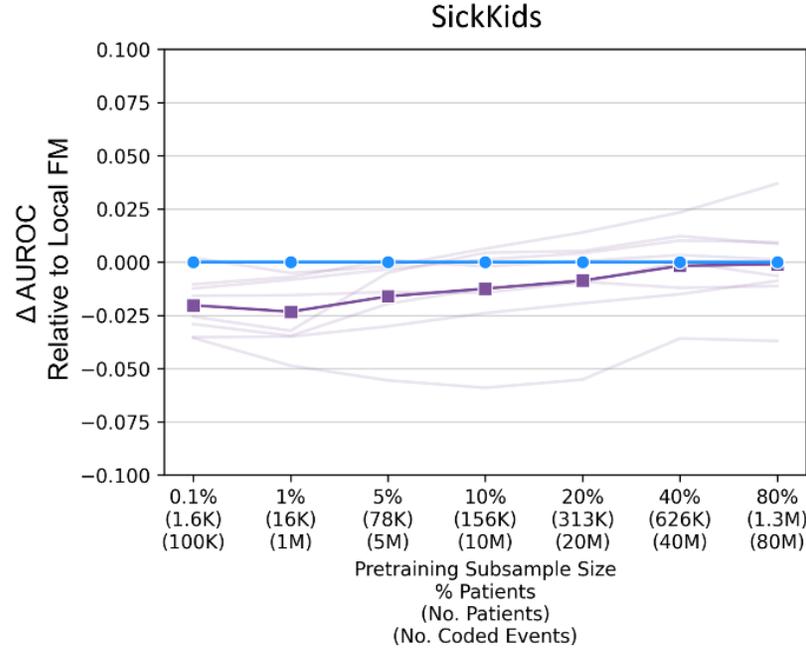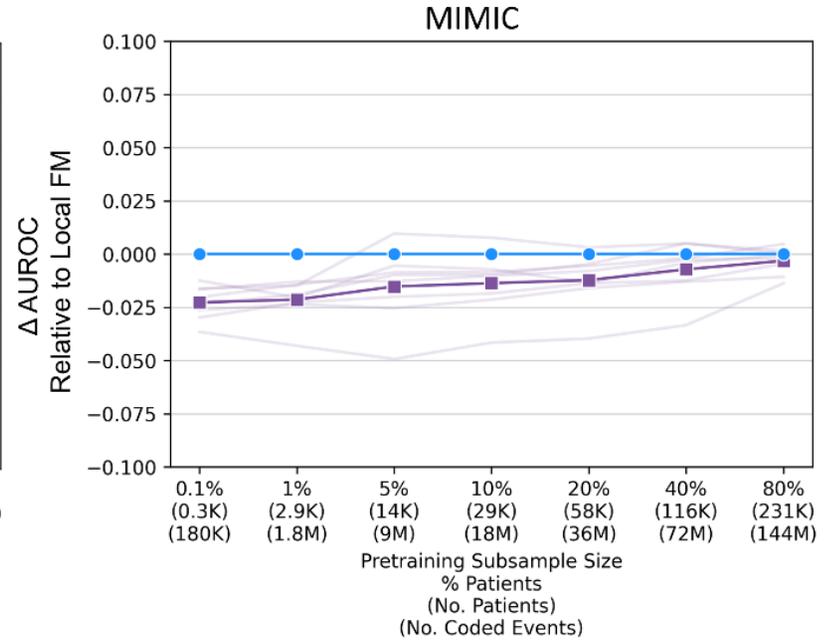

Figure 5. Discrimination of $FM^+_{SM}$ (external foundation model Stanford Medicine with continued pretraining) using increasing pretraining sample size relative to discrimination of local foundation models ($FM_{SK}$ and $FM_{MIMIC}$) pretrained using all data. Bolded and faint lines indicate average and task-specific performance relative to local foundation models ($FM_{SK}$ and $FM_{MIMIC}$), respectively.

Abbreviations: AUROC: area under the receiver operating characteristic curve; $FM^+_{SM}$: external foundation model Stanford Medicine with continued pretraining - SK or MIMIC; $FM_{SK}$/$FM_{MIMIC}$: local foundation model – SK or MIMIC; SK: SickKids; MIMIC: Medical Information Mart for Intensive Care

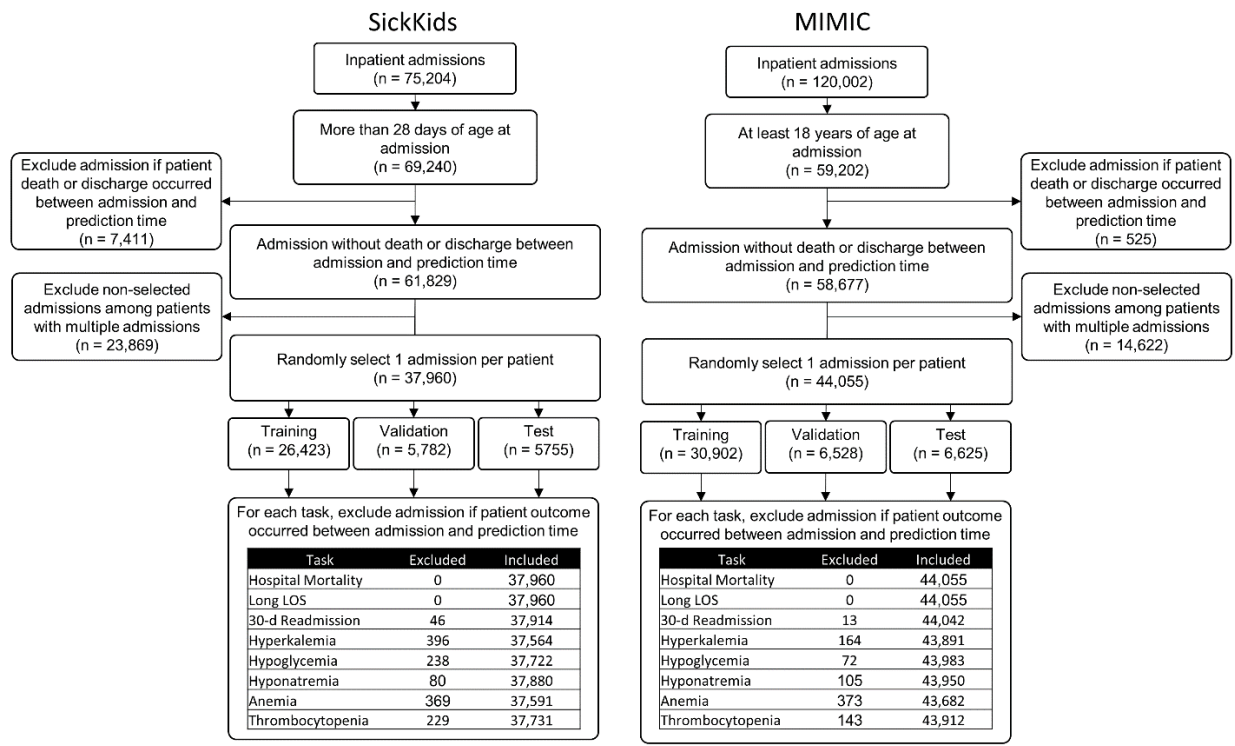

Supplementary Figure 1. Flow diagram of patient selection, inclusion, reason for exclusion and assignment into training, validation, and test sets for task-specific models for each dataset. Patients in the test set were excluded from all pretraining of foundation models.
Abbreviation: LOS: length of stay

Supplementary Table 1. Count-based featurization

| |
|---|
| This conventional featurization approach transforms each patient's electronic health records timeline into a count vector. Each patient's timeline is a sequence of clinical events $X = (x_1, x_2, \ldots, x_n)$, where each $x_i$ denoting the $i$-th code, encompassing any form of structured data obtained from the patient's EHR including diagnosis, lab test, or medication as examples. Each element in the resulting count vector represents the frequency of a specific clinical event (e.g. a diagnosis) occurring within a defined time window (e.g. 1 to 7 days) preceding the prediction time. Featurization was carried out over 3 time windows with respect to prediction time: 0-1 day, 1-7 days, and 7 days to the entire history. The count vectors for all patients were then combined into a high-dimensional (~30K for SK patients, and ~80K for MIMIC patients), sparse count matrix. |

Supplementary Table 2. Hyperparameter settings for each model

| Hyperparameter | Values |
|---|---|
| GBM | |
|    lr | 0.01, 0.1, 0.2 |
|    num_leaves | 100, 300 |
|    boosting_type | "gbdt", "dart", "goss" |
|    n_estimators | 1000 |
| L2-regularized logistic regression | |
|    C | $10^x$ where x ranges from -5 to 1 in 20 equal steps |
|    max_iter | 10,000 |
| *Continued Pretraining and Pretraining of Local Foundation Models | |
|    Learning rate | 1e-4, 1e-5, 1e-6, 1e-7, 1-e8 |

*All foundation models utilized the same transformer architecture with 12 identical layers where each layer is consisted of 12 attention heads and 2 feedforward layers with a local attention mechanism.

Supplementary Table 3. Comparing discrimination and calibration of external vs. local foundation models at each site*

| Model | Discrimination Evaluation | | | Calibration Evaluation | | |
|---|---|---|---|---|---|---|
| | Mean AUROC | Difference [External FM – Local FM] | P-value** | Mean ECE | Difference [External FM – Local FM] | P-value** |
| **SickKids** | | | | | | |
| $FM_{SK}$ | 0.900 [0.849, 0.942] | | | 0.006 [0.003, 0.009] | | |
| $FM_{SM}$ | 0.880 [0.826, 0.928] | -0.019 [-0.035, -0.005] | **0.008** | 0.005 [0.003, 0.009] | 0 [-0.002, 0.002] | 0.816 |
| $FM_{SM}^{+SK}$ | 0.901 [0.851, 0.944] | 0.002 [-0.012, 0.017] | 0.774 | 0.006 [0.003, 0.009] | 0 [-0.001, 0.002] | 0.848 |
| | | | | | | |
| **MIMIC** | | | | | | |
| $FM_{MIMIC}$ | 0.850 [0.793, 0.898] | | | 0.006 [0.004, 0.01] | | |
| $FM_{SM}$ | 0.828 [0.775, 0.875] | -0.023 [-0.035, -0.006] | **0.014** | 0.007 [0.004, 0.012] | 0 [-0.001, 0.004] | 0.761 |
| $FM_{SM}^{+MIMIC}$ | 0.848 [0.783, 0.895] | -0.003 [-0.016, 0.007] | 0.522 | 0.005 [0.003, 0.009] | -0.001 [-0.003, 0.001] | 0.326 |

\* Table shows mean AUROC and ECE across tasks [95% hierarchical bootstrap CI]
\*\* Bolded values indicate P < 0.05
Abbreviations: AUROC: area under the receiver operating characteristics curve; ECE: expected calibration error; CI: confidence interval; $FM_{SM}$: external foundation model Stanford Medicine; $FM_{SM}^{+}$: external foundation model Stanford Medicine with continued pretraining - SK or MIMIC; $FM_{SK}$ or $FM_{MIMIC}$: local foundation model – SK or MIMIC; SK: SickKids; MIMIC: Medical Information Mart for Intensive Care.

Supplementary Table 4. Calibration for task-specific models at two sites*

|  | In-hospital Mortality | Long LOS | 30-day Readmission | Hypoglycemia | Hyponatremia | Hyperkalemia | Thrombocytopenia | Anemia |
|---|---|---|---|---|---|---|---|---|
| **Dataset: SickKids** | | | | | | | | |
| $GBM_{SK}$ | 0.013 [0.012, 0.015] | 0.015 [0.011, 0.025] | 0.014 [0.010, 0.02] | 0.015 [0.013, 0.018] | 0.013 [0.012, 0.015] | 0.011 [0.009, 0.014] | 0.018 [0.016, 0.020] | 0.017 [0.015, 0.021] |
| $FM_{MIMIC}$ | 0.002 [0.001, 0.005] | 0.011 [0.008, 0.021] | 0.007 [0.006, 0.014] | **0.003 [0.002, 0.007]** | 0.002 [0.001, 0.003] | **0.005 [0.004, 0.009]** | 0.002 [0.002, 0.006] | 0.004 [0.003, 0.008] |
| $FM_{MIMIC}^{+SK}$ | **0.001 [0.001, 0.003]** | 0.007 [0.007, 0.019] | 0.009 [0.007, 0.016] | 0.004 [0.002, 0.007] | **0.001 [0.001, 0.002]** | 0.005 [0.003, 0.009] | 0.002 [0.001, 0.005] | 0.005 [0.003, 0.009] |
| $FM_{SM}$ | 0.002 [0.001, 0.004] | 0.011 [0.009, 0.022] | **0.004 [0.004, 0.012]** | 0.004 [0.002, 0.007] | **0.001 [0.001, 0.002]** | 0.005 [0.003, 0.008] | **0.001 [0.001, 0.005]** | **0.003 [0.003, 0.008]** |
| $FM_{SM}^{+SK}$ | 0.002 [0.001, 0.004] | 0.01 [0.008, 0.021] | 0.007 [0.006, 0.014] | 0.004 [0.002, 0.007] | **0.001 [0.001, 0.002]** | 0.005 [0.003, 0.008] | **0.001 [0.001, 0.005]** | 0.005 [0.004, 0.009] |
| $FM_{SK}$ | 0.002 [0.001, 0.004] | **0.006 [0.006, 0.017]** | 0.006 [0.006, 0.015] | 0.005 [0.003, 0.008] | **0.001 [0.001, 0.002]** | 0.006 [0.003, 0.009] | 0.002 [0.001, 0.005] | 0.005 [0.004, 0.01] |
| **Dataset: MIMIC** | | | | | | | | |
| $GBM_{MIMIC}$ | 0.014 [0.013, 0.019] | 0.017 [0.011, 0.027] | 0.015 [0.013, 0.016] | 0.014 [0.011, 0.017] | 0.017 [0.014, 0.019] | 0.017 [0.015, 0.019] | 0.019 [0.016, 0.022] | 0.015 [0.011, 0.021] |
| $FM_{SK}$ | 0.004 [0.003, 0.009] | 0.029 [0.023, 0.040] | 0.004 [0.003, 0.007] | 0.003 [0.003, 0.007] | 0.003 [0.002, 0.005] | 0.002 [0.002, 0.005] | **0.002 [0.002, 0.007]** | 0.008 [0.006, 0.016] |
| $FM_{SK}^{+MIMIC}$ | 0.004 [0.002, 0.009] | 0.022 [0.015, 0.032] | **0.002 [0.002, 0.005]** | 0.004 [0.003, 0.008] | **0.002 [0.002, 0.004]** | 0.003 [0.002, 0.006] | 0.004 [0.003, 0.007] | **0.004 [0.004, 0.011]** |
| $FM_{SM}$ | 0.003 [0.003, 0.008] | 0.02 [0.016, 0.032] | **0.002 [0.002, 0.004]** | 0.004 [0.003, 0.007] | **0.002 [0.002, 0.005]** | 0.003 [0.002, 0.005] | **0.002 [0.002, 0.007]** | 0.007 [0.006, 0.014] |
| $FM_{SM}^{+MIMIC}$ | **0.002 [0.002, 0.007]** | 0.009 [0.008, 0.022] | 0.002 [0.002, 0.005] | 0.003 [0.002, 0.006] | 0.003 [0.002, 0.005] | **0.002 [0.002, 0.005]** | 0.003 [0.003, 0.007] | 0.005 [0.004, 0.012] |
| $FM_{MIMIC}$ | 0.006 [0.003, 0.010] | 0.011 [0.009, 0.023] | **0.002 [0.002, 0.005]** | 0.002 [0.002, 0.006] | 0.003 [0.002, 0.005] | **0.002 [0.002, 0.004]** | 0.004 [0.003, 0.007] | 0.008 [0.006, 0.015] |

*Table shows ECE (95% bootstrap CI) for each task; bolded values indicate lowest ECE across models

Abbreviations: ECE: expected calibration error; GBM: gradient boosting machines; $FM_{SM}$/ $FM_{SK}$/$FM_{MIMIC}$: foundation model Stanford Medicine, SickKids, or MIMIC; $FM_{SM}^{+}$/$FM_{SK}^{+}$/$FM_{MIMIC}^{+}$: external foundation model with continued pretraining - SK or MIMIC; SK: SickKids; MIMIC: Medical Information Mart for Intensive Care; LOS: length of stay; CI: confidence interval

Supplementary Table 5. Comparing discrimination of foundation models vs. baseline GBM using decreasing training samples*

| Training Samples | Foundation Model | Foundation Model Performance | GBM Performance (Baseline) | Difference [Foundation Model – GBM] | P-value ** |
|---|---|---|---|---|---|
| Dataset: SickKids | | | | | |
| 2 | $FM_{SM}$ | 0.598 [0.561, 0.631] | 0.53 [0.512, 0.555] | 0.066 [0.037, 0.096] | **< 0.001** |
| 4 | $FM_{SM}$ | 0.652 [0.606, 0.689] | 0.53 [0.517, 0.546] | 0.121 [0.084, 0.151] | **< 0.001** |
| 8 | $FM_{SM}$ | 0.691 [0.627, 0.742] | 0.573 [0.553, 0.598] | 0.114 [0.067, 0.160] | **< 0.001** |
| 16 | $FM_{SM}$ | 0.738 [0.674, 0.793] | 0.575 [0.547, 0.61] | 0.161 [0.097, 0.216] | **< 0.001** |
| 32 | $FM_{SM}$ | 0.77 [0.711, 0.830] | 0.586 [0.558, 0.609] | 0.185 [0.127, 0.239] | **< 0.001** |
| 64 | $FM_{SM}$ | 0.798 [0.737, 0.854] | 0.704 [0.666, 0.748] | 0.093 [0.055, 0.127] | **< 0.001** |
| 128 | $FM_{SM}$ | 0.82 [0.760, 0.870] | 0.752 [0.713, 0.795] | 0.066 [0.039, 0.096] | **< 0.001** |
| 256 | $FM_{SM}$ | 0.829 [0.773, 0.876] | 0.777 [0.735, 0.820] | 0.049 [0.032, 0.074] | **< 0.001** |
| 512 | $FM_{SM}$ | 0.83 [0.779, 0.876] | 0.792 [0.749, 0.835] | 0.037 [0.019, 0.057] | **< 0.001** |
| 1024 | $FM_{SM}$ | 0.835 [0.787, 0.878] | 0.809 [0.768, 0.850] | 0.025 [0.012, 0.040] | **< 0.001** |
| 2 | $FM_{SM}^{+SK}$ | 0.621 [0.581, 0.661] | 0.53 [0.512, 0.555] | 0.089 [0.061, 0.120] | **< 0.001** |
| 4 | $FM_{SM}^{+SK}$ | 0.674 [0.614, 0.726] | 0.53 [0.517, 0.546] | 0.142 [0.092, 0.190] | **< 0.001** |
| 8 | $FM_{SM}^{+SK}$ | 0.719 [0.660, 0.776] | 0.573 [0.553, 0.598] | 0.144 [0.101, 0.190] | **< 0.001** |
| 16 | $FM_{SM}^{+SK}$ | 0.77 [0.698, 0.830] | 0.575 [0.547, 0.610] | 0.193 [0.122, 0.253] | **< 0.001** |
| 32 | $FM_{SM}^{+SK}$ | 0.807 [0.745, 0.865] | 0.586 [0.558, 0.609] | 0.222 [0.161, 0.275] | **< 0.001** |
| 64 | $FM_{SM}^{+SK}$ | 0.833 [0.771, 0.885] | 0.704 [0.666, 0.748] | 0.127 [0.091, 0.160] | **< 0.001** |
| 128 | $FM_{SM}^{+SK}$ | 0.85 [0.791, 0.899] | 0.752 [0.713, 0.795] | 0.095 [0.068, 0.123] | **< 0.001** |
| 256 | $FM_{SM}^{+SK}$ | 0.858 [0.804, 0.904] | 0.777 [0.735, 0.820] | 0.078 [0.061, 0.100] | **< 0.001** |
| 512 | $FM_{SM}^{+SK}$ | 0.858 [0.807, 0.902] | 0.792 [0.749, 0.835] | 0.064 [0.05, 0.081] | **< 0.001** |
| 1024 | $FM_{SM}^{+SK}$ | 0.86 [0.809, 0.904] | 0.809 [0.768, 0.850] | 0.049 [0.036, 0.065] | **< 0.001** |
| Dataset: MIMIC | | | | | |
| 2 | $FM_{SM}$ | 0.593 [0.569, 0.626] | 0.543 [0.524, 0.562] | 0.051 [0.033, 0.076] | **< 0.001** |
| 4 | $FM_{SM}$ | 0.621 [0.583, 0.664] | 0.529 [0.510, 0.550] | 0.092 [0.068, 0.118] | **< 0.001** |
| 8 | $FM_{SM}$ | 0.651 [0.611, 0.700] | 0.561 [0.538, 0.583] | 0.089 [0.070, 0.121] | **< 0.001** |
| 16 | $FM_{SM}$ | 0.676 [0.639, 0.720] | 0.559 [0.541, 0.575] | 0.117 [0.090, 0.149] | **< 0.001** |
| 32 | $FM_{SM}$ | 0.725 [0.678, 0.771] | 0.567 [0.553, 0.579] | 0.158 [0.107, 0.208] | **< 0.001** |
| 64 | $FM_{SM}$ | 0.745 [0.703, 0.798] | 0.648 [0.617, 0.683] | 0.098 [0.072, 0.122] | **< 0.001** |
| 128 | $FM_{SM}$ | 0.758 [0.709, 0.816] | 0.71 [0.658, 0.777] | 0.048 [0.035, 0.060] | **< 0.001** |
| 256 | $FM_{SM}$ | 0.766 [0.712, 0.828] | 0.746 [0.684, 0.815] | 0.02 [0.006, 0.034] | **0.002** |
| 512 | $FM_{SM}$ | 0.768 [0.710, 0.833] | 0.766 [0.698, 0.835] | 0.003 [-0.007, 0.015] | 0.582 |
| 1024 | $FM_{SM}$ | 0.768 [0.708, 0.835] | 0.775 [0.706, 0.842] | -0.007 [-0.020, 0.008] | 0.375 |
| 2 | $FM_{SM}^{+MIMIC}$ | 0.612 [0.579, 0.650] | 0.543 [0.524, 0.562] | 0.069 [0.048, 0.092] | **< 0.001** |
| 4 | $FM_{SM}^{+MIMIC}$ | 0.658 [0.608, 0.700] | 0.529 [0.510, 0.550] | 0.128 [0.090, 0.162] | **< 0.001** |
| 8 | $FM_{SM}^{+MIMIC}$ | 0.704 [0.647, 0.761] | 0.561 [0.538, 0.583] | 0.143 [0.105, 0.181] | **< 0.001** |
| 16 | $FM_{SM}^{+MIMIC}$ | 0.739 [0.678, 0.798] | 0.559 [0.541, 0.575] | 0.18 [0.128, 0.228] | **< 0.001** |
| 32 | $FM_{SM}^{+MIMIC}$ | 0.765 [0.702, 0.820] | 0.567 [0.553, 0.579] | 0.198 [0.133, 0.258] | **< 0.001** |
| 64 | $FM_{SM}^{+MIMIC}$ | 0.785 [0.725, 0.842] | 0.648 [0.617, 0.683] | 0.136 [0.099, 0.169] | **< 0.001** |
| 128 | $FM_{SM}^{+MIMIC}$ | 0.798 [0.74, 0.858] | 0.71 [0.658, 0.775] | 0.087 [0.071, 0.104] | **< 0.001** |
| 256 | $FM_{SM}^{+MIMIC}$ | 0.801 [0.743, 0.863] | 0.746 [0.684, 0.817] | 0.055 [0.041, 0.069] | **< 0.001** |
| 512 | $FM_{SM}^{+MIMIC}$ | 0.803 [0.745, 0.866] | 0.766 [0.698, 0.835] | 0.038 [0.026, 0.050] | **< 0.001** |
| 1024 | $FM_{SM}^{+MIMIC}$ | 0.807 [0.749, 0.869] | 0.775 [0.706, 0.842] | 0.033 [0.022, 0.046] | **< 0.001** |

\* Table shows mean AUROC (95% hierarchical bootstrap CI) by number of training samples

\*\* bolded values indicate P < 0.05

Abbreviations: AUROC: area under the receiver operating characteristics curve. GBM: gradient boosting machines; $FM_{SM}$: external foundation model Stanford Medicine; $FM_{SM}^{+}$: external FM Stanford Medicine with continued pretraining - SK or MIMIC; SK: SickKids; MIMIC: Medical Information Mart for Intensive Care; CI – confidence interval.

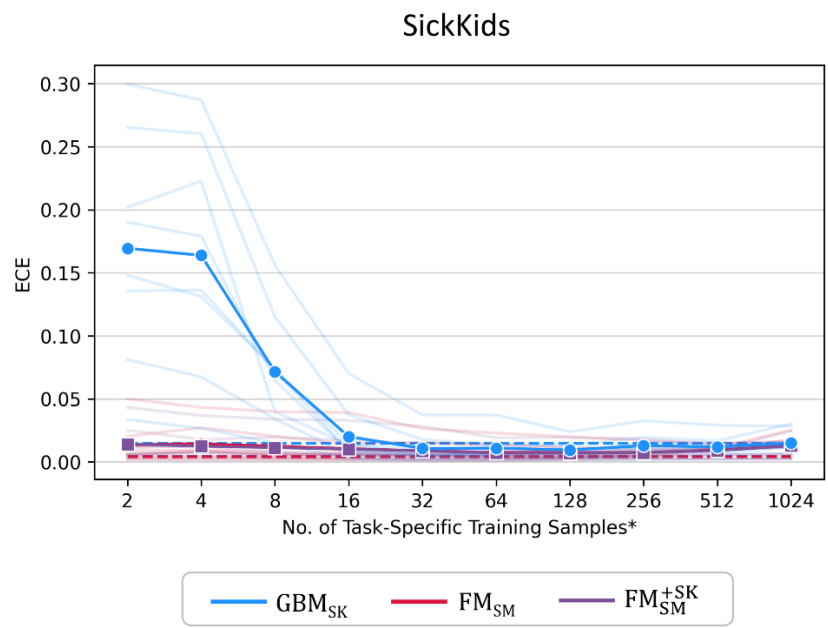 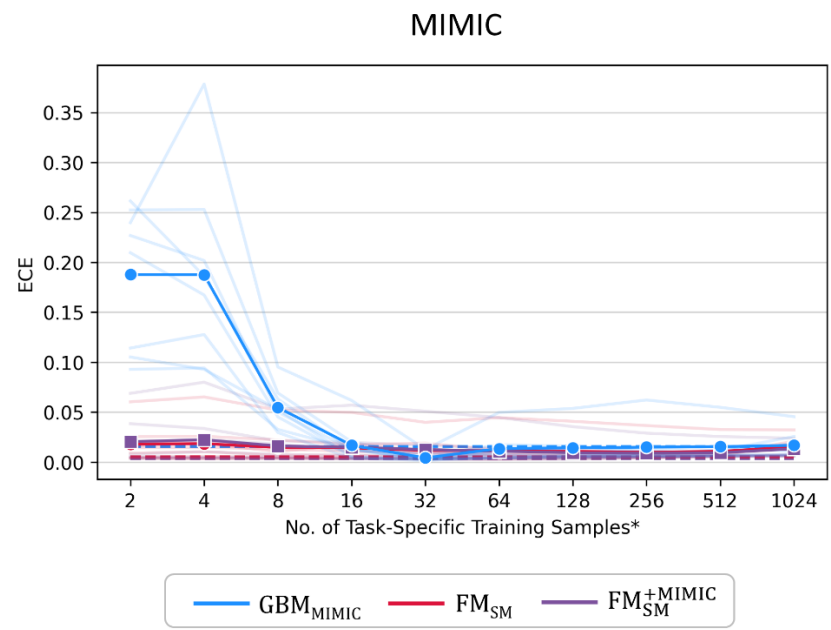

Supplementary Figure 2. Calibration of external foundation model ($FM_{SM}$), external foundation model with continued pretraining ($FM_{SM}^{+}$) and baseline GBM using decreasing training samples at SickKids and MIMIC. Bolded and faint lines indicate average and task-specific performance, respectively. Dashed lines indicate mean ECE of models trained on all training samples.

* The number of training examples for each class is up to half of the number of task-specific training samples.

Abbreviations: ECE: estimated calibration error; $FM_{SM}$: external foundation model Stanford Medicine; $FM_{SM}^{+}$: external foundation model Stanford Medicine with continued pretraining – SK or MIMIC; SK: SickKids; MIMIC: Medical Information Mart for Intensive Care; GBM: gradient boosting machines.

Supplementary Table 6. Comparing calibration of foundation models vs. baseline GBM using decreasing task-specific training samples*

| Training Samples | Foundation Model | Foundation Model Performance | GBM Performance (Baseline) | Difference [Foundation Model – GBM] | P-value ** |
|---|---|---|---|---|---|
| Dataset: SickKids | | | | | |
| 2 | $FM_{SM}$ | 0.014 [0.006, 0.026] | 0.169 [0.112, 0.225] | -0.155 [-0.202, -0.103] | **< 0.001** |
| 4 | $FM_{SM}$ | 0.015 [0.007, 0.026] | 0.162 [0.107, 0.223] | -0.148 [-0.198, -0.099] | **< 0.001** |
| 8 | $FM_{SM}$ | 0.013 [0.006, 0.022] | 0.071 [0.045, 0.102] | -0.058 [-0.081, -0.038] | **< 0.001** |
| 16 | $FM_{SM}$ | 0.011 [0.005, 0.020] | 0.020 [0.008, 0.036] | -0.009 [-0.016, -0.003] | **< 0.001** |
| 32 | $FM_{SM}$ | 0.010 [0.005, 0.016] | 0.011 [0.005, 0.020] | -0.001 [-0.004, 0.000] | 0.102 |
| 64 | $FM_{SM}$ | 0.008 [0.004, 0.014] | 0.011 [0.006, 0.020] | -0.003 [-0.007, -0.001] | **< 0.001** |
| 128 | $FM_{SM}$ | 0.008 [0.004, 0.013] | 0.010 [0.006, 0.015] | -0.002 [-0.003, -0.001] | **< 0.001** |
| 256 | $FM_{SM}$ | 0.008 [0.005, 0.012] | 0.013 [0.007, 0.020] | -0.005 [-0.008, -0.002] | **< 0.001** |
| 512 | $FM_{SM}$ | 0.010 [0.007, 0.013] | 0.012 [0.007, 0.018] | -0.002 [-0.005, 0.000] | 0.1 |
| 1024 | $FM_{SM}$ | 0.014 [0.009, 0.018] | 0.015 [0.009, 0.022] | -0.001 [-0.005, 0.001] | 0.24 |
| 2 | $FM_{SM}^{+SK}$ | 0.015 [0.007, 0.025] | 0.169 [0.112, 0.225] | -0.154 [-0.202, -0.102] | **< 0.001** |
| 4 | $FM_{SM}^{+SK}$ | 0.013 [0.007, 0.022] | 0.162 [0.107, 0.223] | -0.149 [-0.201, -0.100] | **< 0.001** |
| 8 | $FM_{SM}^{+SK}$ | 0.012 [0.006, 0.020] | 0.071 [0.045, 0.102] | -0.059 [-0.083, -0.038] | **< 0.001** |
| 16 | $FM_{SM}^{+SK}$ | 0.011 [0.005, 0.019] | 0.020 [0.008, 0.036] | -0.009 [-0.018, -0.002] | **< 0.001** |
| 32 | $FM_{SM}^{+SK}$ | 0.009 [0.004, 0.016] | 0.011 [0.005, 0.020] | -0.002 [-0.004, -0.000] | **0.008** |
| 64 | $FM_{SM}^{+SK}$ | 0.008 [0.004, 0.013] | 0.011 [0.006, 0.020] | -0.004 [-0.008, -0.001] | **0.004** |
| 128 | $FM_{SM}^{+SK}$ | 0.007 [0.004, 0.012] | 0.01 0[0.006, 0.015] | -0.003 [-0.004, -0.001] | **< 0.001** |
| 256 | $FM_{SM}^{+SK}$ | 0.008 [0.005, 0.012] | 0.013 [0.007, 0.020] | -0.005 [-0.009, -0.002] | **< 0.001** |
| 512 | $FM_{SM}^{+SK}$ | 0.010 [0.007, 0.012] | 0.012 [0.007, 0.018] | -0.003 [-0.006, 0.001] | 0.132 |
| 1024 | $FM_{SM}^{+SK}$ | 0.013 [0.009, 0.018] | 0.015 [0.009, 0.022] | -0.002 [-0.006, 0.001] | 0.204 |
| Dataset: MIMIC | | | | | |
| 2 | $FM_{SM}$ | 0.018 [0.008, 0.033] | 0.188 [0.138, 0.232] | -0.169 [-0.209, -0.128] | **< 0.001** |
| 4 | $FM_{SM}$ | 0.018 [0.009, 0.034] | 0.184 [0.135, 0.254] | -0.165 [-0.22, -0.125] | **< 0.001** |
| 8 | $FM_{SM}$ | 0.015 [0.007, 0.028] | 0.055 [0.042, 0.070] | -0.040 [-0.045, -0.034] | **< 0.001** |
| 16 | $FM_{SM}$ | 0.015 [0.007, 0.027] | 0.017 [0.008, 0.032] | -0.002 [-0.006, 0.002] | 0.404 |
| 32 | $FM_{SM}$ | 0.012 [0.006, 0.022] | 0.004 [0.003, 0.007] | 0.008 [0.003, 0.015] | **< 0.001** |
| 64 | $FM_{SM}$ | 0.012 [0.006, 0.023] | 0.014 [0.007, 0.026] | -0.002 [-0.004, -0.001] | **< 0.001** |
| 128 | $FM_{SM}$ | 0.011 [0.006, 0.021] | 0.015 [0.007, 0.028] | -0.003 [-0.007, -0.001] | **< 0.001** |
| 256 | $FM_{SM}$ | 0.010 [0.006, 0.019] | 0.015 [0.007, 0.030] | -0.005 [-0.011, -0.001] | **< 0.001** |
| 512 | $FM_{SM}$ | 0.011 [0.007, 0.019] | 0.015 [0.008, 0.028] | -0.004 [-0.010, -0.001] | **< 0.001** |
| 1024 | $FM_{SM}$ | 0.015 [0.010, 0.022] | 0.017 [0.009, 0.027] | -0.002 [-0.006, 0.002] | 0.426 |
| 2 | $FM_{SM}^{+MIMIC}$ | 0.020 [0.009, 0.038] | 0.188 [0.138, 0.232] | -0.166 [-0.206, -0.127] | **< 0.001** |
| 4 | $FM_{SM}^{+MIMIC}$ | 0.021 [0.010, 0.040] | 0.184 [0.135, 0.254] | -0.162 [-0.214, -0.124] | **< 0.001** |
| 8 | $FM_{SM}^{+MIMIC}$ | 0.017 [0.008, 0.030] | 0.055 [0.042, 0.070] | -0.038 [-0.044, -0.032] | **< 0.001** |
| 16 | $FM_{SM}^{+MIMIC}$ | 0.015 [0.006, 0.030] | 0.017 [0.008, 0.032] | -0.001 [-0.004, 0.001] | 0.214 |
| 32 | $FM_{SM}^{+MIMIC}$ | 0.014 [0.006, 0.025] | 0.004 [0.003, 0.007] | 0.009 [0.003, 0.018] | **< 0.001** |

| 64 | $FM_{SM}^{+MIMIC}$ | 0.011 [0.005, 0.023] | 0.014 [0.007, 0.026] | -0.003 [-0.004, -0.001] | **< 0.001** |
| 128 | $FM_{SM}^{+MIMIC}$ | 0.010 [0.005, 0.019] | 0.015 [0.007, 0.028] | -0.005 [-0.009, -0.002] | **< 0.001** |
| 256 | $FM_{SM}^{+MIMIC}$ | 0.009 [0.005, 0.017] | 0.015 [0.007, 0.030] | -0.006 [-0.014, -0.001] | **< 0.001** |
| 512 | $FM_{SM}^{+MIMIC}$ | 0.010 [0.007, 0.016] | 0.015 [0.008, 0.028] | -0.005 [-0.013, -0.001] | **< 0.001** |
| 1024 | $FM_{SM}^{+MIMIC}$ | 0.014 [0.009, 0.018] | 0.017 [0.009, 0.027] | -0.003 [-0.009, 0.001] | 0.193 |

\* Table shows mean ECE (95% hierarchical bootstrap CI) by number of training samples

\*\* bolded values indicate P < 0.05

Abbreviations: ECE: expected calibration error. GBM: gradient boosting machines; $FM_{SM}$: external foundation model Stanford Medicine; $FM_{SM}^{+}$: external foundation model Stanford Medicine with continued pretraining - SK or MIMIC; SK: SickKids; MIMIC: Medical Information Mart for Intensive Care; CI – confidence interval

Supplementary Table 7. Comparing discrimination of external vs. local foundation models approaches at each pretraining sample size*

| Proportion Pretraining Samples | External Foundation Model | External Foundation Model Performance | Local Foundation Model Performance | Difference [External – Local] | P-value |
|---|---|---|---|---|---|
| SickKids | | | | | |
| 0.001 | $FM_{SM}$ | 0.88 [0.826, 0.928] | 0.794 [0.741, 0.843] | -0.086 [-0.112, -0.062] | **< 0.001** |
| 0.01 | $FM_{SM}$ | 0.88 [0.826, 0.928] | 0.837 [0.788, 0.887] | -0.043 [-0.071, -0.02] | **< 0.001** |
| 0.05 | $FM_{SM}$ | 0.88 [0.826, 0.928] | 0.866 [0.821, 0.91] | -0.014 [-0.033, 0.005] | 0.142 |
| 0.1 | $FM_{SM}$ | 0.88 [0.826, 0.928] | 0.875 [0.831, 0.914] | -0.006 [-0.021, 0.011] | 0.48 |
| 0.2 | $FM_{SM}$ | 0.88 [0.826, 0.928] | 0.891 [0.84, 0.937] | 0.011 [-0.001, 0.023] | 0.078 |
| 0.4 | $FM_{SM}$ | 0.88 [0.826, 0.928] | 0.891 [0.84, 0.936] | 0.011 [-0.01, 0.025] | 0.23 |
| 0.8 | $FM_{SM}$ | 0.88 [0.826, 0.928] | 0.9 [0.851, 0.941] | 0.019 [0.006, 0.034] | **0.002** |
| 0.001 | $FM_{SM}^{+SK}$ | 0.879 [0.828, 0.922] | 0.794 [0.741, 0.843] | -0.085 [-0.111, -0.061] | **< 0.001** |
| 0.01 | $FM_{SM}^{+SK}$ | 0.876 [0.824, 0.922] | 0.837 [0.787, 0.887] | -0.039 [-0.065, -0.016] | **< 0.001** |
| 0.05 | $FM_{SM}^{+SK}$ | 0.884 [0.828, 0.933] | 0.866 [0.821, 0.91] | -0.017 [-0.035, 0.002] | 0.078 |
| 0.1 | $FM_{SM}^{+SK}$ | 0.887 [0.833, 0.936] | 0.875 [0.831, 0.914] | -0.013 [-0.03, 0.005] | 0.166 |
| 0.2 | $FM_{SM}^{+SK}$ | 0.891 [0.836, 0.938] | 0.891 [0.84, 0.937] | -0.0 [-0.011, 0.013] | 0.964 |
| 0.4 | $FM_{SM}^{+SK}$ | 0.898 [0.845, 0.944] | 0.891 [0.84, 0.936] | -0.006 [-0.028, 0.005] | 0.366 |
| 0.8 | $FM_{SM}^{+SK}$ | 0.899 [0.85, 0.941] | 0.9 [0.851, 0.941] | 0.0 [-0.014, 0.016] | 0.956 |
| MIMIC | | | | | |
| 0.001 | $FM_{SM}$ | 0.828 [0.775, 0.875] | 0.762 [0.705, 0.812] | -0.066 [-0.089, -0.048] | **< 0.001** |
| 0.01 | $FM_{SM}$ | 0.828 [0.775, 0.875] | 0.815 [0.775, 0.858] | -0.013 [-0.03, 0.013] | 0.253 |
| 0.05 | $FM_{SM}$ | 0.828 [0.775, 0.875] | 0.833 [0.787, 0.88] | 0.006 [-0.008, 0.023] | 0.352 |
| 0.1 | $FM_{SM}$ | 0.828 [0.775, 0.875] | 0.836 [0.783, 0.884] | 0.01 [-0.006, 0.024] | 0.218 |
| 0.2 | $FM_{SM}$ | 0.828 [0.775, 0.875] | 0.84 [0.785, 0.889] | 0.013 [-0.002, 0.025] | 0.073 |
| 0.4 | $FM_{SM}$ | 0.828 [0.775, 0.875] | 0.845 [0.794, 0.892] | 0.017 [0.003, 0.031] | **0.028** |
| 0.8 | $FM_{SM}$ | 0.828 [0.775, 0.875] | 0.845 [0.774, 0.898] | 0.018 [-0.01, 0.036] | 0.183 |
| 0.001 | $FM_{SM}^{+MIMIC}$ | 0.828 [0.765, 0.88] | 0.761 [0.705, 0.812] | -0.066 [-0.088, -0.045] | **< 0.001** |
| 0.01 | $FM_{SM}^{+MIMIC}$ | 0.83 [0.766, 0.882] | 0.815 [0.775, 0.858] | -0.015 [-0.036, 0.021] | 0.39 |
| 0.05 | $FM_{SM}^{+MIMIC}$ | 0.837 [0.768, 0.887] | 0.833 [0.787, 0.88] | -0.003 [-0.021, 0.031] | 0.849 |
| 0.1 | $FM_{SM}^{+MIMIC}$ | 0.838 [0.77, 0.889] | 0.836 [0.783, 0.884] | -0.0 [-0.017, 0.024] | 0.991 |
| 0.2 | $FM_{SM}^{+MIMIC}$ | 0.839 [0.769, 0.89] | 0.84 [0.785, 0.889] | 0.002 [-0.01, 0.02] | 0.759 |
| 0.4 | $FM_{SM}^{+MIMIC}$ | 0.844 [0.775, 0.894] | 0.845 [0.794, 0.892] | 0.001 [-0.012, 0.026] | 0.925 |
| 0.8 | $FM_{SM}^{+MIMIC}$ | 0.848 [0.782, 0.896] | 0.845 [0.774, 0.898] | -0.002 [-0.021, 0.011] | 0.804 |

* Table shows mean AUROC (95% hierarchical bootstrap CI) by proportion of pretraining cohort size
** Bolded values indicate P<0.05.
Abbreviations: AUROC: area under the receiver operating characteristics curve. $FM_{SM}$: external foundation model Stanford Medicine; $FM_{SM}^{+}$: external foundation model Stanford Medicine with continued pretraining - SK or MIMIC; SK: SickKids; MIMIC: Medical Information Mart for Intensive Care; CI: confidence interval.

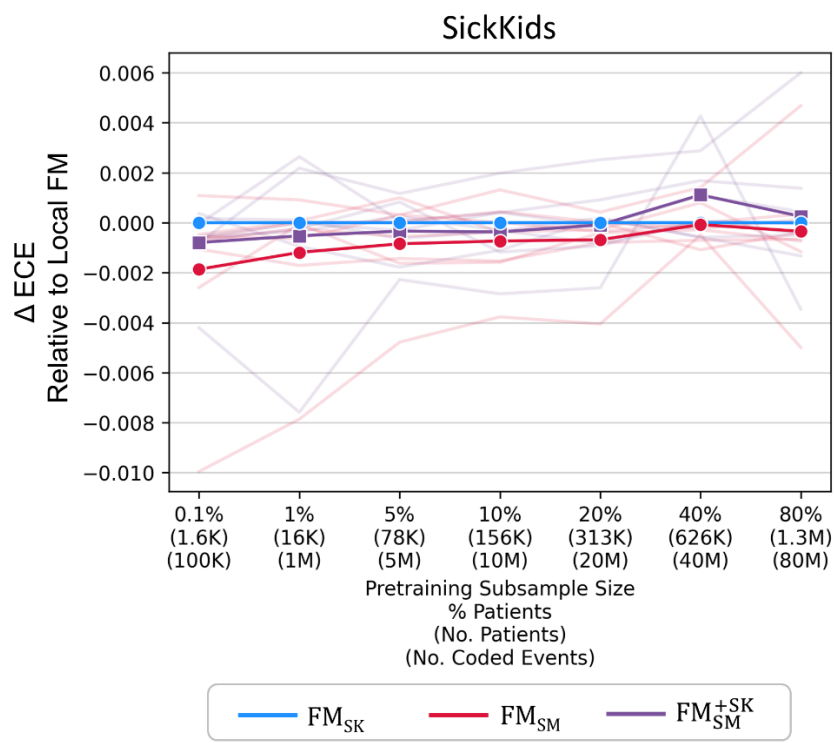 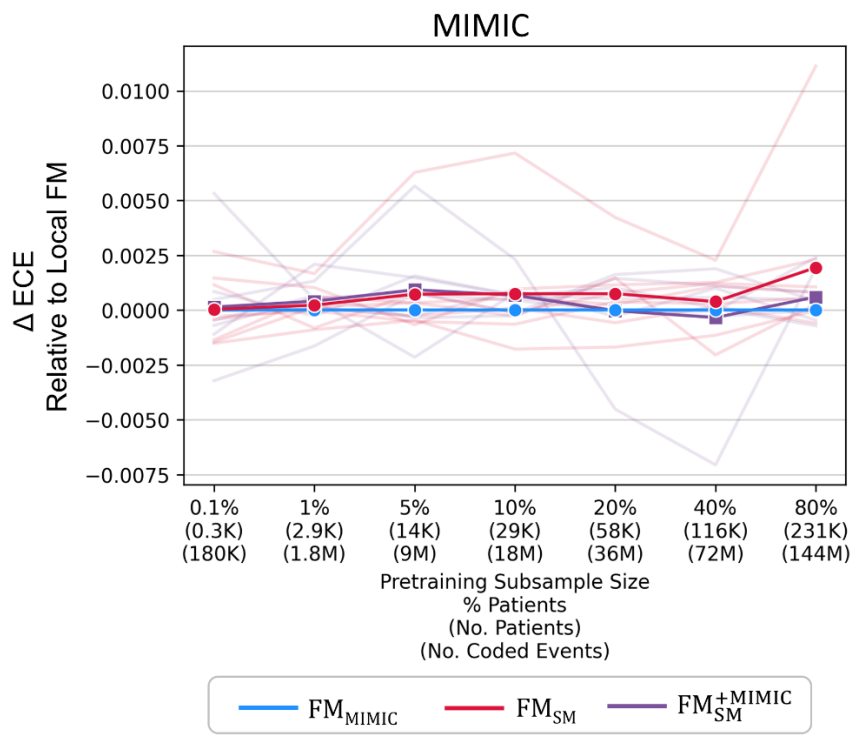

Supplementary Figure 3. Calibration of external foundation model ($FM_{SM}$) and external foundation model with continued pretraining ($FM_{SM}^{+}$) relative to local foundation model ($FM_{SK}$ and $FM_{MIMIC}$) using decreasing pretraining sample size.

Bolded and faint lines indicate average and task-specific performance relative to local foundation models, respectively. The subsample size is not relevant for $FM_{SM}$, which did not undergo additional pretraining. Note, the model ECE scores are relative to the baseline local models ($FM_{SK}$ and $FM_{MIMIC}$) and the absolute ECE does change across sample sizes.

Abbreviations: ECE: expected calibration error; $FM_{SM}$: external foundation model Stanford Medicine; $FM_{SM}^{+}$: external foundation model Stanford Medicine with continued pretraining – SK or MIMIC; $FM_{SK}$/$FM_{MIMIC}$: local foundation model – SK or MIMIC; SK: SickKids; MIMIC: Medical Information Mart for Intensive Care.

Supplementary Table 8. Comparing calibration of external vs. local foundation models at each pretraining sample size*

| Proportion Pretraining Samples | External Foundation Model | External Foundation Model Performance | Local Foundation Model Performance | Difference [External – Local] | P-value |
|---|---|---|---|---|---|
| SickKids | | | | | |
| 0.001 | $FM_{SM}$ | 0.005 [0.003, 0.009] | 0.007 [0.004, 0.011] | 0.002 [-0.001, 0.004] | 0.162 |
| 0.01 | $FM_{SM}$ | 0.005 [0.003, 0.009] | 0.007 [0.004, 0.01] | 0.001 [-0.001, 0.003] | 0.2 |
| 0.05 | $FM_{SM}$ | 0.005 [0.003, 0.009] | 0.006 [0.003, 0.01] | 0.001 [-0.001, 0.002] | 0.414 |
| 0.1 | $FM_{SM}$ | 0.005 [0.003, 0.009] | 0.006 [0.003, 0.009] | 0.0 [-0.001, 0.002] | 0.602 |
| 0.2 | $FM_{SM}$ | 0.005 [0.003, 0.009] | 0.006 [0.003, 0.009] | 0.0 [-0.002, 0.002] | 0.788 |
| 0.4 | $FM_{SM}$ | 0.005 [0.003, 0.009] | 0.005 [0.003, 0.009] | -0.0 [-0.002, 0.001] | 0.924 |
| 0.8 | $FM_{SM}$ | 0.005 [0.003, 0.009] | 0.005 [0.003, 0.009] | 0.0 [-0.002, 0.002] | 0.94 |
| 0.001 | $FM_{SM}^{+SK}$ | 0.006 [0.003, 0.009] | 0.007 [0.004, 0.011] | 0.001 [-0.001, 0.003] | 0.226 |
| 0.01 | $FM_{SM}^{+SK}$ | 0.006 [0.003, 0.009] | 0.007 [0.004, 0.01] | 0.001 [-0.001, 0.003] | 0.518 |
| 0.05 | $FM_{SM}^{+SK}$ | 0.006 [0.003, 0.01] | 0.006 [0.003, 0.01] | 0.0 [-0.001, 0.002] | 0.64 |
| 0.1 | $FM_{SM}^{+SK}$ | 0.005 [0.003, 0.009] | 0.006 [0.003, 0.009] | 0.0 [-0.001, 0.002] | 0.594 |
| 0.2 | $FM_{SM}^{+SK}$ | 0.006 [0.003, 0.01] | 0.006 [0.003, 0.009] | 0.0 [-0.002, 0.002] | 0.994 |
| 0.4 | $FM_{SM}^{+SK}$ | 0.006 [0.003, 0.01] | 0.005 [0.003, 0.009] | -0.001 [-0.002, 0.001] | 0.374 |
| 0.8 | $FM_{SM}^{+SK}$ | 0.006 [0.003, 0.01] | 0.005 [0.003, 0.009] | -0.0 [-0.002, 0.001] | 0.924 |
| MIMIC | | | | | |
| 0.001 | $FM_{SM}$ | 0.007 [0.004, 0.012] | 0.007 [0.004, 0.012] | 0.0 [-0.002, 0.002] | 0.737 |
| 0.01 | $FM_{SM}$ | 0.007 [0.004, 0.012] | 0.007 [0.004, 0.012] | -0.0 [-0.002, 0.002] | 0.992 |
| 0.05 | $FM_{SM}$ | 0.007 [0.004, 0.012] | 0.006 [0.003, 0.01] | -0.001 [-0.003, 0.001] | 0.414 |
| 0.1 | $FM_{SM}$ | 0.007 [0.004, 0.012] | 0.006 [0.004, 0.01] | -0.001 [-0.003, 0.001] | 0.516 |
| 0.2 | $FM_{SM}$ | 0.007 [0.004, 0.012] | 0.006 [0.003, 0.01] | -0.001 [-0.003, 0.001] | 0.432 |
| 0.4 | $FM_{SM}$ | 0.007 [0.004, 0.012] | 0.006 [0.003, 0.011] | -0.001 [-0.003, 0.001] | 0.524 |
| 0.8 | $FM_{SM}$ | 0.007 [0.004, 0.012] | 0.005 [0.003, 0.009] | -0.001 [-0.004, 0.0] | 0.093 |
| 0.001 | $FM_{SM}^{+MIMIC}$ | 0.007 [0.003, 0.012] | 0.007 [0.004, 0.012] | 0.0 [-0.002, 0.002] | 0.876 |
| 0.01 | $FM_{SM}^{+MIMIC}$ | 0.007 [0.004, 0.012] | 0.007 [0.004, 0.012] | 0.0 [-0.002, 0.002] | 0.998 |
| 0.05 | $FM_{SM}^{+MIMIC}$ | 0.007 [0.004, 0.011] | 0.006 [0.003, 0.01] | -0.0 [-0.002, 0.001] | 0.517 |
| 0.1 | $FM_{SM}^{+MIMIC}$ | 0.006 [0.004, 0.011] | 0.006 [0.004, 0.01] | -0.0 [-0.002, 0.001] | 0.619 |
| 0.2 | $FM_{SM}^{+MIMIC}$ | 0.006 [0.004, 0.009] | 0.006 [0.003, 0.01] | 0.0 [-0.001, 0.002] | 0.961 |
| 0.4 | $FM_{SM}^{+MIMIC}$ | 0.006 [0.004, 0.01] | 0.006 [0.003, 0.011] | 0.0 [-0.002, 0.002] | 0.939 |
| 0.8 | $FM_{SM}^{+MIMIC}$ | 0.006 [0.003, 0.01] | 0.005 [0.003, 0.009] | -0.0 [-0.002, 0.001] | 0.542 |

* Table shows mean ECE (95% hierarchical bootstrap CI) by proportion of pretraining cohort size
** Bolded values indicate P<0.05.
Abbreviations: ECE: expected calibration error. $FM_{SM}$: external foundation model Stanford Medicine; $FM_{SM}^{+}$: external foundation model Stanford Medicine with continued pretraining - SK or MIMIC; SK: SickKids; MIMIC: Medical Information Mart for Intensive Care; CI: confidence interval.

Supplementary Table 9. Comparing discrimination of local foundation models pretrained on all data with continued pretraining of $FM_{SM}$ using increasing pretraining sample size*

| Proportion Pretraining Samples | $FM_{SM}^+$ | Local FM ($FM_{SK}$ or $FM_{MIMIC}$) | Difference [External – Local] | P-value |
|---|---|---|---|---|
| SickKids | | | | |
| 0.001 | 0.879 [0.828, 0.922] | 0.900 [0.849, 0.942] | -0.020 [-0.033, -0.008] | **0.004** |
| 0.01 | 0.876 [0.824, 0.922] | 0.900 [0.849, 0.942] | -0.023 [-0.038, -0.010] | **< 0.001** |
| 0.05 | 0.884 [0.828, 0.933] | 0.900 [0.849, 0.942] | -0.016 [-0.032, -0.001] | **0.038** |
| 0.1 | 0.887 [0.833, 0.936] | 0.900 [0.849, 0.942] | -0.012 [-0.030, 0.004] | 0.138 |
| 0.2 | 0.891 [0.836, 0.938] | 0.900 [0.849, 0.942] | -0.008 [-0.027, 0.007] | 0.306 |
| 0.4 | 0.898 [0.845, 0.944] | 0.900 [0.849, 0.942] | -0.001 [-0.016, 0.013] | 0.850 |
| 0.8 | 0.899 [0.850, 0.941] | 0.900 [0.849, 0.942] | -0.001 [-0.017, 0.017] | 0.932 |
| MIMIC | | | | |
| 0.001 | 0.828 [0.765, 0.880] | 0.850 [0.792, 0.898] | -0.022 [-0.038, -0.010] | **<0.001** |
| 0.01 | 0.830 [0.766, 0.881] | 0.850 [0.793, 0.898] | -0.020 [-0.037, -0.010] | **<0.001** |
| 0.05 | 0.837 [0.768, 0.887] | 0.850 [0.792, 0.898] | -0.014 [-0.032, -0.001] | **0.037** |
| 0.1 | 0.838 [0.770, 0.889] | 0.850 [0.792, 0.898] | -0.013 [-0.030, -0.001] | **0.042** |
| 0.2 | 0.839 [0.769, 0.890] | 0.850 [0.792, 0.898] | -0.011 [-0.028, -0.001] | **0.046** |
| 0.4 | 0.844 [0.775, 0.894] | 0.850 [0.792, 0.898] | -0.006 [-0.024, 0.005] | 0.316 |
| 0.8 | 0.848 [0.782, 0.896] | 0.850 [0.792, 0.898] | -0.003 [-0.016, 0.008] | 0.575 |

* Table shows mean AUROC (95% hierarchical bootstrap CI) by proportion of pretraining cohort size

** Bolded values indicate P<0.05.

Abbreviations: AUROC: area under the receiver operating characteristics curve; $FM_{SM}^+$: external foundation model Stanford Medicine with continued pretraining - SK or MIMIC; $FM_{SK}$/$FM_{MIMIC}$: local foundation model – SK or MIMIC; SK: SickKids; MIMIC: Medical Information Mart for Intensive Care; CI: confidence interval.

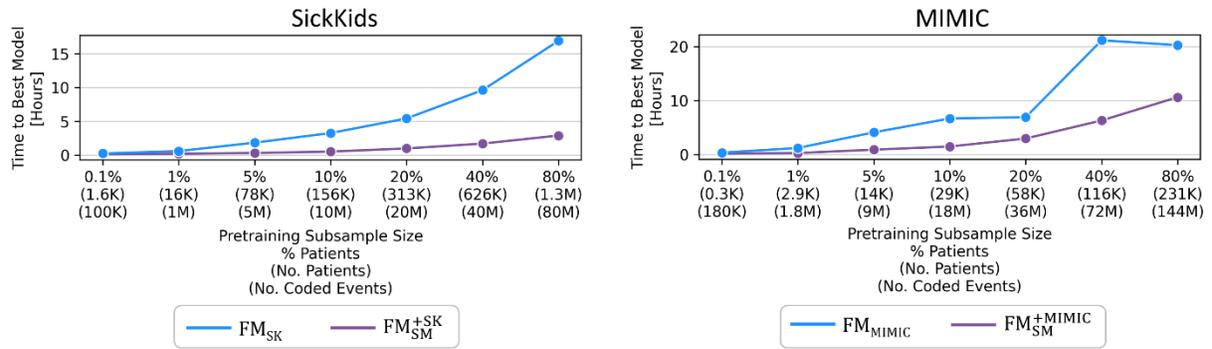

Supplementary Figure 4. Pretraining time for continued pretraining compared to local foundation models ($FM_{SK}$ and $FM_{MIMIC}$) measured as the average time to best model across hyperparameter settings for each pretraining sample size.

Abbreviations: $FM_{SM}^{+}$: external foundation model Stanford Medicine with continued pretraining - SK or MIMIC; $FM_{SK}$/$FM_{MIMIC}$: local foundation model – SK or MIMIC; SK: SickKids; MIMIC: Medical Information Mart for Intensive Care.

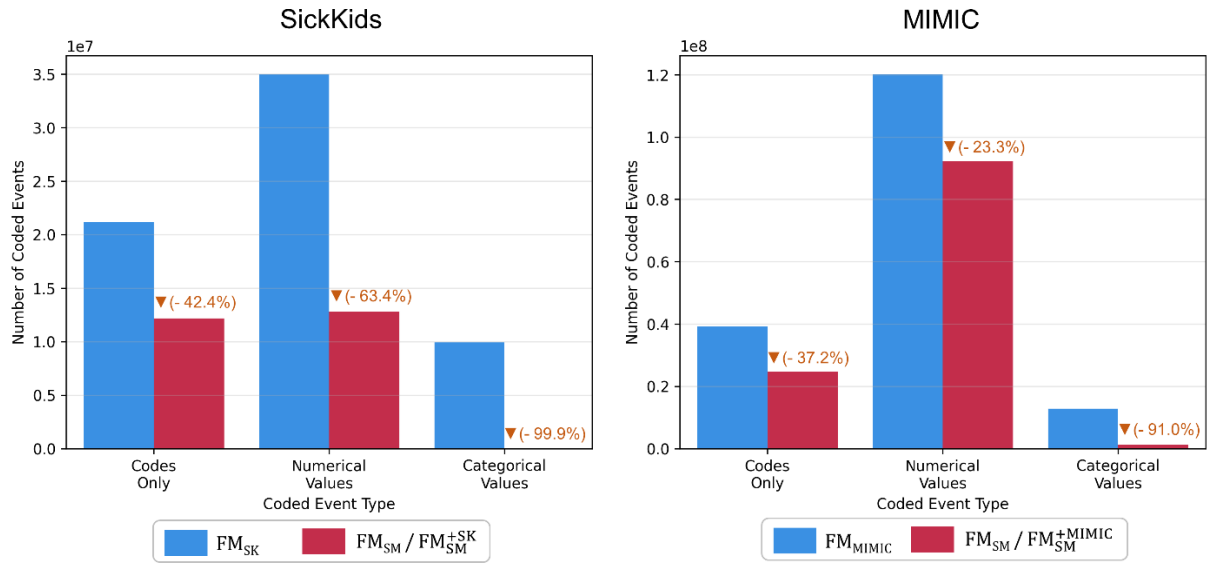

Supplementary Figure 5. Number of SickKids/MIMIC coded events across all patient timelines processed by local (blue) and external (red) foundation models, with values in orange indicating the relative percentage of coded events not processed (due to not being supported) by external foundation models. This involved an independent analysis in which we used each foundation model to process the entire timeline of each patient in the dataset.

Abbreviations: $FM_{SM}$: external foundation model Stanford Medicine; $FM_{SM}^{+}$: external foundation model Stanford Medicine with continued pretraining - SK or MIMIC; $FM_{SK}/FM_{MIMIC}$: local foundation model – SK or MIMIC; SK: SickKids; MIMIC: Medical Information Mart for Intensive Care.